\documentclass[letterpaper]{article} 
\usepackage{st_template}  
\usepackage{times}  
\usepackage{helvet}  
\usepackage{courier}  
\usepackage[hyphens]{url}  
\usepackage{graphicx} 
\usepackage{natbib}  
\usepackage{caption} 
\usepackage{algorithm}
\usepackage{algorithmic}

\usepackage{amsmath, amssymb}
\usepackage{microtype}
\usepackage{url}
\usepackage{booktabs}
\usepackage{lineno}
\usepackage{subcaption}
\usepackage{longtable}
\usepackage{multirow}
\usepackage{xcolor,colortbl}

\definecolor{g}{HTML}{DBF9CC}
\definecolor{g2}{HTML}{B0F893}
\definecolor{r}{HTML}{FCE2D5}
\definecolor{infoseek}{HTML}{78206E}
\definecolor{ours}{HTML}{215F9A}

\newcommand{\ds}{\textsc{Visual-RAG }}
\newcommand{\dsns}{\textsc{Visual-RAG}}
\newcommand{\cc}[1]{\cellcolor{#1}}

\usepackage{newfloat}
\usepackage{listings}
\DeclareCaptionStyle{ruled}{labelfont=normalfont,labelsep=colon,strut=off} 
\floatstyle{ruled}
\newfloat{listing}{tb}{lst}{}
\floatname{listing}{Listing}
\title{Visual-RAG: Benchmarking Text-to-Image Retrieval Augmented Generation for Visual Knowledge Intensive Queries}
\author{
    Yin Wu\textsuperscript{\rm 1}, Quanyu Long\textsuperscript{\rm 1}, Jing Li\textsuperscript{\rm 2}, Jianfei Yu\textsuperscript{\rm 3}, Wenya Wang\textsuperscript{\rm 1}\\
}
\affiliations{
    \textsuperscript{\rm 1}Nanyang Technological University,\\
    \textsuperscript{\rm 2}Harbin Institute of Technology (Shenzhen),
    \textsuperscript{\rm 3}Nanjing University of Science and Technology\\

    wuyi0023@e.ntu.edu.sg
}

\begin{document}

\maketitle

\begin{abstract}
Retrieval‑augmented generation (RAG) is a paradigm that augments large language models (LLMs) with external knowledge to tackle knowledge‑intensive question answering.
While several benchmarks evaluate Multimodal LLMs (MLLMs) under Multimodal RAG settings, they predominantly retrieve from textual corpora and do not explicitly assess how models exploit visual evidence during generation. Consequently, there still lacks benchmark that isolates and measures the contribution of retrieved images in RAG.
We introduce \dsns, a question‑answering benchmark that targets visually grounded, knowledge‑intensive questions. Unlike prior work, \ds requires text‑to‑image retrieval and the integration of retrieved clue images to extract visual knowledge necessary for answer generation.
With \dsns, we evaluate 5 open-source and 3 proprietary MLLMs and demonstrate that images provide strong evidence in augmented generation; however, even state‑of‑the‑art models struggle to effectively extract and utilize visual knowledge. Our results highlight the need for improved visual retrieval, grounding, and attribution in multimodal RAG systems.\footnote{Benchmark available at: \url{github.com/visual-rag/visual-rag}}
\end{abstract}



\section{Introduction}
\label{sec:intro}
\begin{figure*}[ht]
    \centering
    \includegraphics[width=\textwidth]{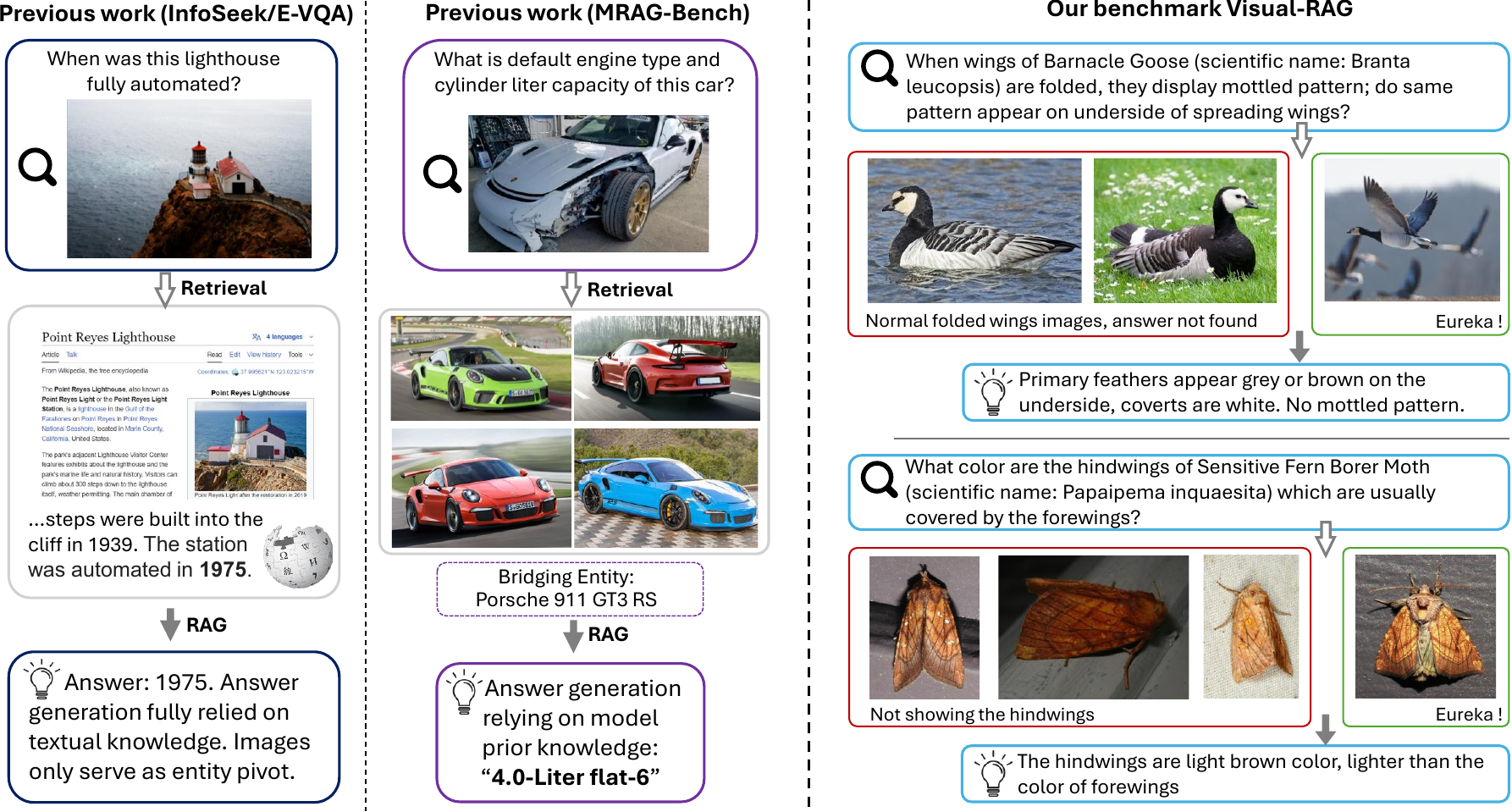}
    \caption{InfoSeek style: Textual-Knowledge-Centric RAG (left), MRAG-Bench style: Retrieving common images for entity recognition and entity bridged prior knowledge recalling (middle) v.s. Ours: Visual Evidence-Centric Retrieval and Visual Knowledge-Intensive Reasoning (right).
    MRAG-Bench~\cite{hu2024mragbenchvisioncentricevaluationretrievalaugmented} formulates its task by presenting uncommon images that are challenging for entity recognition, retrieving more common images to support the MLLMs in bridging object identification. Typically, the actual answers (e.g., identifying the engine used in a specific car) is rarely explicitly represented visually in the retrieved images, making the answer dependent heavily on the model’s internal prior knowledge.
    In contrast, our \ds benchmark utilizes text-only queries with a common identifier (e.g., name of an organism), and requires retrieval of specialized, less common images containing explicit answer within their visual contents. Consequently, \ds makes cross-modal retrieval more challenging and demands deeper visually grounded knowledge acquisition, thus facilitating a more precise visual-centric evaluation of MLLMs' capability to extract detailed visual information for knowledge-intensive reasoning.}
    \label{fig:question_sample}
\end{figure*}
Retrieval‑Augmented Generation (RAG) augments large language models (LLMs) with external knowledge, substantially improving performance on knowledge‑intensive queries~\citep{NEURIPS2020_6b493230, karpukhin-etal-2020-dense, izacard-grave-2021-leveraging}. While the application of RAG is well-acknowledged for text-only knowledge-intensive tasks, many real‑world scenarios require multimodal knowledge from visual cues, such as diagrams or maps, highlighting the limitations of relying solely on text retrieval. Several benchmarks have been proposed to evaluate Multimodal Large Language Models (MLLMs) under multimodal RAG settings, targeting complex, knowledge‑intensive questions. Notably, InfoSeek~\cite{chen-etal-2023-pre-trained} and Encyclopedic-VQA~\cite{Mensink_2023_ICCV} pair mixed‑modal queries directly with images. Yet, these benchmarks are fundamentally \emph{Textual‑Knowledge‑Centric}: answering typically involves first recognizing entities in the query image and then retrieving textual evidence (e.g., Wikipedia passages) to generate the answer, as demonstrated in Figure~\ref{fig:question_sample} (left). As a result, although images are present in query, MLLMs rarely need to extract visual knowledge from them during answer generation, and current multimodal RAG benchmarks predominantly assess textual rather than visual knowledge exploitation.

Recent work MRAG-Bench~\cite{hu2024mragbenchvisioncentricevaluationretrievalaugmented} partially addresses this gap by explicitly requiring image retrieval. As shown in Figure~\ref{fig:question_sample} (middle), MRAG-Bench features challenging images as queries (e.g., damaged, partially occluded or incomplete entities), enforcing models to retrieve more common images of the entity to assist entity recognition. However, final answers (e.g., the engine used in a specific car model) largely rely on parametric model knowledge rather than explicit visual evidence contained in the retrieved images. Consequently, solving the task still depends on internal priors instead of information extracted from the pictures, leaving the core challenge of visually grounded RAG unaddressed. 

We therefore introduce \dsns, a visual knowledge‑intensive QA benchmark that explicitly requires \textbf{text‑to‑image retrieval} and \textbf{visually grounded answer generation}. Crucially, the retrieved image itself serves as evidence, which aligns with real-world use case: once a clue image containing the relevant visual feature is surfaced, one can answer the visual knowledge-intensive question directly, without treating the image as a pivot to trigger subsequent recalling of prior knowledge. Accordingly, \ds emphasizes \textit{Visual Evidence‑Centric Retrieval} and mandates \textit{Visual Knowledge‑Intensive Reasoning}, challenging models to extract and ground the information contained in clue images, and to discriminate it from visually similar hard-negative distractors. Key features of \ds include:

\begin{enumerate}
    \item Fine‑grained and less common visual evidence. Only 6.12\% of corpus images are clue images containing the necessary visual evidence, substantially making cross‑modal retrieval challenging.
    \item Hard negatives at scale. Most non‑clue images are hard negatives, which depict the same entities and are visually similar, stress‑testing both retrieval models' robustness and downstream MLLMs' capabilities in visual knowledge extraction.
    \item Explicit visual answerability. Evidences of answers are encoded directly in the visual content, enabling evaluation of whether MLLMs truly leverage visual information.
\end{enumerate}

To construct \dsns, we initially generate candidate queries using LLMs, followed by rigorous refinement through human filtering and rewriting processes. Subsequently, we employ both open-source MLLMs and human annotators for additional filtering, ensuring the final queries reflect genuine human curiosity and practical relevance.

\newcommand{\iconimg}{\includegraphics[width=1.2em]{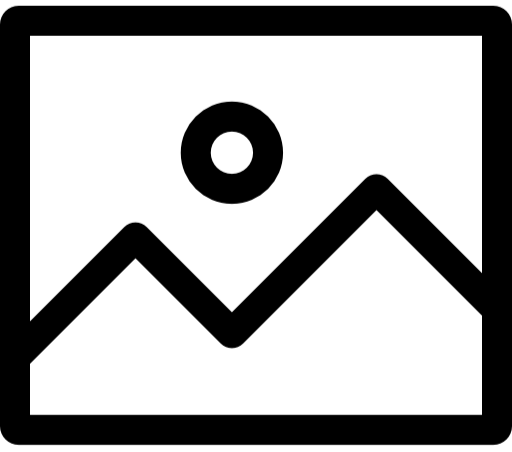}}
\newcommand{\icontext}{\includegraphics[width=1.3em, height=1.07em]{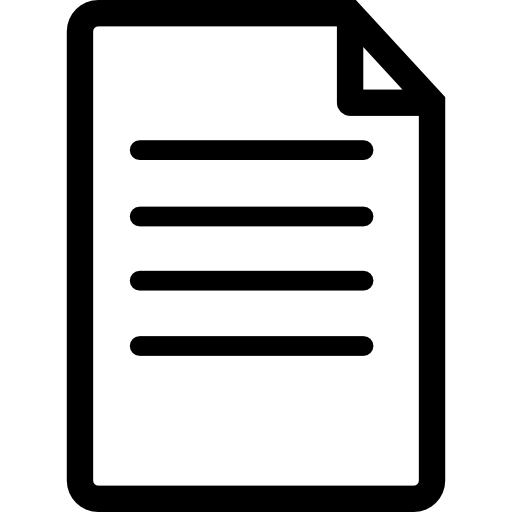}}
\newcommand{\iconarrow}{\includegraphics[width=1.1em]{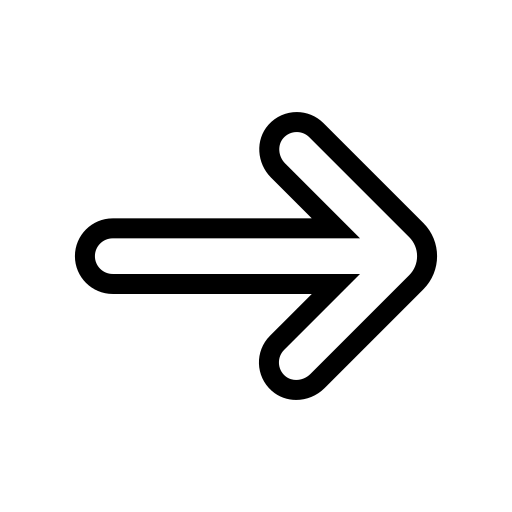}}
\newcommand{\iconlink}{\includegraphics[width=1em]{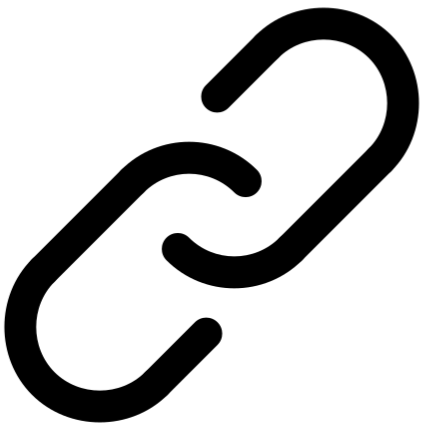}}

\begin{table*}[ht]
    \centering
    \begin{tabular}{l|ccc}
    \toprule
         Benchmark  &Query Modality   &Knowledge Modality  &Retrieval Modality\\
        \midrule
         InfoSeek~\cite{chen-etal-2023-pre-trained}   &\raisebox{-2pt}{\iconimg\icontext} &\raisebox{-2pt}{\icontext}  &\raisebox{-2pt}{\iconimg  \iconarrow \iconimg \hspace{1pt}\iconlink  \icontext \iconarrow \icontext}\\
         E-VQA~\cite{Mensink_2023_ICCV}  &\raisebox{-2pt}{\iconimg\icontext} &\raisebox{-2pt}{\icontext}  &\raisebox{-2pt}{\iconimg  \iconarrow \iconimg \hspace{1pt}\iconlink\icontext \iconarrow \icontext}\\
         WebQA~\cite{Chang_2022_CVPR}  &\raisebox{-2pt}{\icontext}  &\raisebox{-2pt}{\iconimg\icontext}  &\raisebox{-2pt}{\icontext \iconarrow \icontext\iconlink \hspace{1pt}\iconimg}\\
         MRAG-Bench~\cite{hu2024mragbenchvisioncentricevaluationretrievalaugmented} &\raisebox{-2pt}{\iconimg\icontext}  &\raisebox{-2pt}{\iconimg}   &\raisebox{-2pt}{\iconimg\icontext \iconarrow \iconimg}\\
         \ds (Ours) &\raisebox{-2pt}{\icontext}  &\raisebox{-2pt}{\iconimg}   &\raisebox{-2pt}{\icontext \iconarrow \iconimg}\\
    \bottomrule
    \end{tabular}
    \caption{Comparing modalities of knowledge-intensive VQA benchmarks. The link icon \raisebox{-2pt}{\iconlink} denotes that the two objects are paired, retrieving one will automatically link to the other. For InfoSeek and E-VQA, \citet{yan-xie-2024-echosight} demonstrated that by image-to-image retrieval on Wikipedia images, the Wikipedia articles containing those images are also retrieved at a significantly higher recall comparing to image-to-text retrieval. Similarly for WebQA, a question-to-caption (text-to-text) retrieval achieves better performance than text-to-image retrieval, as shown in Appendix~\ref{sec:app_relation}. Our \ds enforces text-to-image retrieval for visual knowledge.} %
    \label{tab:comapre_bench}
\end{table*}

With \dsns, we comprehensively evaluate eight popular MLLMs, including five open-source and three proprietary models. Our baseline results confirm that images can serve as strong evidence for augmented generation. We also examine multiple RAG configurations to assess how effectively MLLMs ingest image augmentation, and to analyze their ability to discriminate informative visual clues from irrelevant distractors. The key findings from our experiments are summarized as follows:
\begin{itemize}
    \item \textbf{Visual evidence-centric retrieval is challenging}. The commonly used cross-modal retrievers face significant challenge in addressing difficult queries requiring the identification of fine-grained visual evidence.
    \item \textbf{Visual clues can serve as effective evidence for Retrieval-Augmented Generation.} The evaluated MLLMs demonstrate the ability to extract visual knowledge to answer questions when presented with single ground-truth clue image. 
    \item \textbf{Realistic visual RAG remains an onerous task.} With top-$k$ RAG, performance of most models lags behind the ground-truth clue upper bound. Proprietary models reach the upper bound only when $k$ close to 20, revealing limited ability to efficiently exploit visual clues, partially constrained by the bottleneck of retriever.
    \item \textbf{Inverted dynamics of open-source and proprietary models.} When a guaranteed clue is mixed with non-clue images, open-source MLLMs leverage it well at lower $k$, but their accuracy decays as including more distractors. In contrast, proprietary MLLMs retain robustness in identifying visual evidence from distractors at higher $k$, yet overlook the clue when only a few images are provided.
\end{itemize}


\section{Related Work}
Numerous benchmarks have been developed to evaluate the capabilities of MLLMs. We focus on those specifically designed for knowledge-intensive visual question answering (VQA), particularly emphasizing retrieval-augmented generation (RAG) paradigms.  Table \ref{tab:comapre_bench} provides an overview and comparison of modalities across widely adopted benchmarks.

\paragraph{Textual Knowledge Centric VQA Benchmarks.}
Benchmarks such as OK-VQA~\cite{Marino_2019_CVPR} and its augmented variant, A-OKVQA~\cite{10.1007/978-3-031-20074-8_9}, predominantly emphasize commonsense and world knowledge but exhibit limited dependence on external knowledge retrieval mechanisms. Other datasets like ViQuAE~\cite{10.1145/3477495.3531753}, InfoSeek~\cite{chen-etal-2023-pre-trained} and Encyclopedic-VQA (E-VQA, \citealp{Mensink_2023_ICCV}) address encyclopedic questions paired with images of specific entities. However, the retrieved information employed in answering is primarily textual (e.g., passages from Wikipedia), with the visual modality mainly serving as an entity anchor rather than as a source of knowledge.

\paragraph{Visual Knowledge Centric QA Benchmarks.}
To date, only a limited number of benchmarks explicitly require visual knowledge retrieval as evidence in answering questions.
WebQA~\cite{Chang_2022_CVPR} is an earlier effort that necessitates retrieving images for visual knowledge but relies heavily on textual captions, as images in the benchmark often lack explicit indicators for being clue.
More recently, MRAG-Bench~\cite{hu2024mragbenchvisioncentricevaluationretrievalaugmented} similarly emphasizes image retrieval. However, MRAG-Bench provides queries in the form of challenging images depicting partially obscured or incomplete entities, aiming primarily at entity recognition through matching visually similar images. The retrieved images predominantly help in identifying entities, with answers typically dependent on the model’s prior knowledge rather than directly on visual evidence within these images.

Unlike MRAG-Bench, our benchmark, \ds, explicitly constructs textual queries and requires retrieving specialized images that explicitly contain visual evidence. Consequently, our benchmark evaluates deeper visual knowledge extraction capabilities for MLLMs, emphasizing visual evidence-centric reasoning. More detailed comparison between \ds and prior benchmarks is provided in Appendix~\ref{sec:app_relation}.


\section{Benchmark Overview}
\label{sec:overview}

We propose \dsns, a Question Answering (QA) benchmark for visual knowledge–intensive factual queries. \ds evaluates two capabilities under retrieval‑augmented generation (RAG): (i) \emph{visual evidence–centric} text‑to‑image retrieval and (ii) an MLLM’s ability for \emph{visual knowledge-intensive reasoning} with retrieved images in retrieval-augmented answer generation.

\paragraph{Task Definition.} 
Let $\mathcal{Q}$ be a set of text queries and $\mathcal{D}$ a corpus of images. For each $q \in \mathcal{Q}$ there exists a small set of \emph{clue images} $\mathcal{I}_c \subset \mathcal{D}$ whose \emph{pixel content} explicitly encodes the visual evidence needed to answer $q$. The remaining images $\mathcal{I}_d = \mathcal{D} \setminus \mathcal{I}_c$ act as distractors; many are \emph{hard negatives} that depict the same entity or highly similar instances but \emph{lack} the queried visual feature.

Given $q$, a system first retrieves $k$ images $\mathcal{I}_{ret;k} \subset \mathcal{D}$ and then generates an answer $\hat{y}$ conditioned on $(q, I_{ret;k})$. Crucially, \ds assumes that once a correct clue image is retrieved, the answer is directly derivable from that image, without requiring subsequent textual lookup.

\paragraph{Visual Knowledge–Intensive Queries.}
Queries in \ds are \emph{text‑only}. They probe factual properties of visual features that generally hold for an entity category (e.g., ``\emph{What color are the hindwings of \textit{Papaipema inquaesita}?}'', as in Figure \ref{fig:question_sample}), rather than instance‑specific properties typical of standard VQA (e.g., ``\emph{What color is the cup on the table?}''). We focus on the organism domain because it naturally supports fine‑grained, specialized questions for which the necessary evidence is visual and stable at the category level. Unlike many existing works, we deliberately exclude image from the query input to avoid reducing the task to entity recognition or image-to-image unimodal similarity matching; the primary objective is cross‑modal knowledge extraction from retrieved images.
Numerous established benchmarks already cover entity recognition, including the OVEN~\cite{Hu_2023_ICCV} dataset which serves as the foundation for the InfoSeek dataset, and the iNaturalist 2021 (\citealp{Van_Horn_2021_CVPR}) dataset, which we employ in this work. Moreover, our setting of visual knowledge-intensive queries simulates realistic application scenarios, where referring directly to images for answers is more intuitive and efficient compared to searching through textual documentation for such query.

\paragraph{Naturally Co-occurring Hard Negative Images.} \ds incorporates a large number of naturally co‑occurring hard negatives: images of the same species that do \emph{not} exhibit the queried feature. All clue images are retrievable based on visual content alone; captions or alt‑text are not required or assumed at retrieval time. This construction jointly pressures the retriever - to surface the rare, feature‑bearing images, and the answer generator MLLM - to extract and ground the relevant visual evidence.

\begin{figure}[t]
\centering
    \begin{subfigure}[b]{0.49\linewidth}
    \centering
        \includegraphics[width=11.7em]{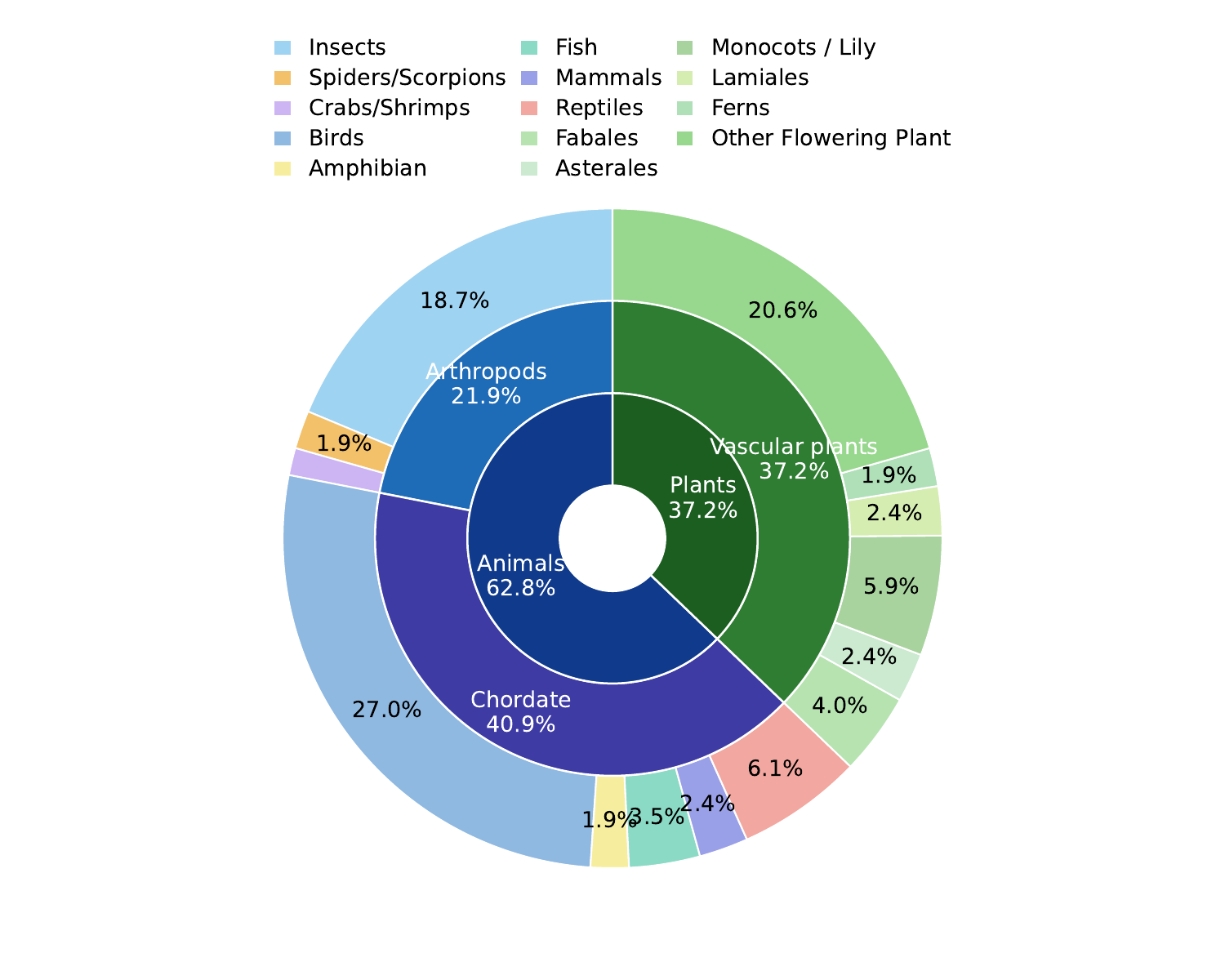}
        \caption{Organism categories.}
    \end{subfigure}
    \hfill
    \begin{subfigure}[b]{0.49\linewidth}
    \centering
        \includegraphics[width=11em]{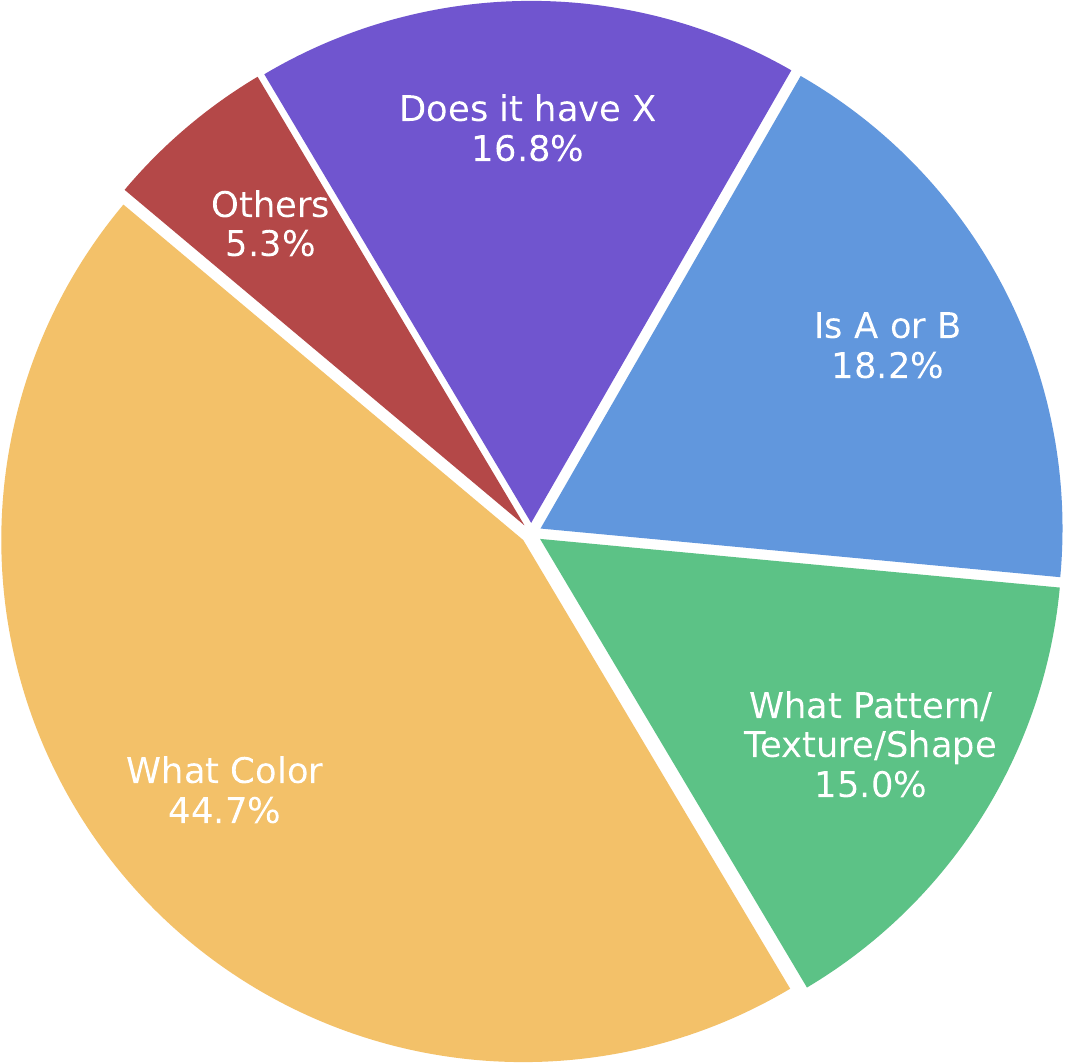}
        \caption{Question categories.}
    \end{subfigure}
    \caption{Distribution of organisms and question categories in \ds.} 
    \label{fig:q_distribution}
\end{figure}

\paragraph{Dataset Statistics.}
Our \ds contains 374 queries, forming an image knowledge base of total 99,017 images. On average, each query has 16.19 clue images; across the corpus, clue images constitute 6.12\% of all images. The distributions of organism categories and query types are shown in Figure~\ref{fig:q_distribution}.

\section{Benchmark Construction}
\label{sec:const}

\subsection{Data source}

Following Encyclopedic VQA~\cite{Mensink_2023_ICCV}, we adopt iNaturalist 2021 (iNat21, \citealp{Van_Horn_2021_CVPR}) as the cornerstone of our benchmark. iNat21 contains 2.7M images of organism (animals, plants, fungi, etc.), spanning across 10000 taxonomy species. Each species in iNat21 has 200–300 images, originally collected for species-level image classification.

\subsection{Query Collection}

\paragraph{Question generation.}
Our goal is to create text queries about distinct visual features of organisms such that, for a given species, only a small fraction of images exhibit the queried feature. Each query must admit a unique answer or answer set that generally holds at the species level (category‑level fact), rather than instance‑specific properties.

Constructing such queries by hand would require domain expertise and careful inspection of roughly 200–300 images per species. To scale this process, we employ OpenAI‑o1 as a query generator. For each candidate species, we provide its Wikipedia Summary and Description sections to the model to convey basic context. We explicitly instruct the model \emph{NOT} to generate questions about properties already described in these passages, pushing it toward specialized visual attributes that are unlikely to appear in standard references. The model outputs (i) a set of candidate questions and (ii) their associated \emph{visual feature descriptors} (e.g., ``hindwing color'', ``presence of mottled pattern'', etc.).

To reduce the chance that answers can be trivially recovered from text corpora, we bias species selection toward those that are relatively under‑represented in Wikipedia. A subsequent refined human filtering and rewriting process (Appendix~\ref{sec:app-anno}, \ref{sec:app_prompt}) validates the linguistic form of each query and the feature descriptors. %


\paragraph{Question Filtering and Image Annotation.}
Our curation process enforces two constraints: (C1) \emph{clue images} (images that clearly display the queried feature with the required view/pose) must be a minority within the species‑specific sub‑corpus; and (C2) each query must be \emph{answerable} by at least one clue image. Formally, for a query $q$ over species $s(q)$ with image set $\mathcal{D}_{s(q)}$, let $\mathcal{I}_c \subset \mathcal{D}_{s(q)}$ denote the clue set and define the \emph{clue rate}
$\rho_q \triangleq \frac{|\mathcal{I}_c|}{|\mathcal{D}_{s(q)}|}$. We retain only queries within threshold $0 < \rho_q \le 0.25$.

\paragraph{Stage 1: MLLM‑based coarse filter.}

We use an open‑source MLLM\footnote{Llama-3.2-11B-Vision-Instruct. Prompt template can be found in Appendix~\ref{sec:app_prompt}.} as an image‑level presence / absence classifier for the referenced feature. For each query $q$, the model iterates over $\mathcal{D}_{s(q)}$ one image at a time and predicts the binary label indicating whether the feature is visually present. From these labels we obtain an estimate clue rate $\hat{\rho}_q$ and discard queries with $\hat{\rho}_q > 0.25$. The premise is that when a queried feature is unusually common, it is sufficiently prominent for the MLLM to detect reliably, making the coarse estimate adequate for pre‑screening.

\paragraph{Stage 2: Human annotation and answer adjudication.}
We recruit volunteer university students as annotators to (a) label images by whether the queried feature is \emph{visibly} present and (b) provide the ground‑truth answer for the query $q$. This step confirms that (i) at least one valid clue image exists and (ii) $\rho_q \le 0.25$, thereby removing queries that slipped past the coarse filter. We take the human labels as ground truth for $\mathcal{I}_c$ and compute the final $\rho_q$ accordingly.
Approximately $1{,}000$ questions pass the coarse MLLM filter. After human verification, 374 questions satisfy the clue‑rate criterion ($0 < \rho_q \le 0.25$) and are confirmed answerable by at least one clue image.


\section{Experiments}
\label{sec:exp}

\subsection{Experimental Setups}
\paragraph{Retrieval Evaluation.}
\label{sec:eval_ret}
For retrieval, we evaluate the performance with Recall score, Normalized Discounted Cumulative Gain (NDCG), Hit rate and Hit Count. 
Recall@$k$ measures the fraction of all ground-truth clue images that appear within the top-$k$ results, reflecting overall coverage.
NDCG@$k$ rewards correct images that appear higher in the ranking by discounting relevance logarithmically with position, thus capturing ranking quality.
Hit@$k$ is a binary success indicator for probability of at least one clue is retrieved in the top-$k$ images, while Hit Count@$k$ tallies how many clue images are found in that cut-off, offering an intuitive sense of how much useful evidence the retriever supplies.

Due to the challenge of complex domain-specific queries, it is significantly difficult for the retrievers to search through the entire iNat21 corpus (2.7M images) while the query is about a sophisticated feature for a single species. To enhance retrieval effectiveness and emphasize the more fine-grained and distinguishing features of a species, we narrow the search space by limiting the retrieval corpus to $\mathcal{D}_{s(q)}$, the set of 200–300 images corresponding to the species for each query.

\paragraph{Dense Retrievers.}
Three dense retriever models are evaluated: (1) The commonly used multimodal retriever CLIP (clip-vit-large-patch14-336, \citealp{pmlr-v139-radford21a}). (2) A recent CLIP variant, BGE (BGE-VL-large, \citealp{zhou2024megapairs}. (3) MLLM-based retriever, VLM2Vec-qwen2VL-2B~\cite{jiang2025vlmvec}. The search indexes are constructed with the FAISS library~\cite{douze2024faisslibrary} using flat inner product index. RAG experiments in following sections adopt images retrieved by CLIP, as CLIP is being adopted as the de-facto retriever in previous multimodal RAG works.




\paragraph{Generation Evaluation.}
\label{sec:eval_gen}
Previous work has relied predominantly on exact-match or recall-based metrics to assess answer correctness. However, such methods are prone to overlooking partial hallucinations blended with ground-truth answers. For example, if the true answer is ``black and white'' but the model predicts ``black and white with yellow dots'', the spurious mention of ``yellow dots'' remains undetected under exact-match criteria.

To address this limitation, we adopt \textbf{LLM-as-Judge}, employing OpenAI o4-mini as a more nuanced evaluator. We retain all annotator-provided answers as valid ground-truth candidates, and specify in instruction prompt to consider alternative expressions of features. Detailed evaluation prompt template can be found in Appendix~\ref{sec:app_prompt}. A series of works, including but not limited to \citet{kamalloo-etal-2023-evaluating,zheng2023judging,huang2024empiricalstudyllmasajudgellm}, have demonstrated that general-purpose LLMs can reliably judge open-ended QA responses. We also manually evaluated 200 sampled predicted answers, and the LLM labels yielded alignment score of $F_1=0.92$ with human annotator. The confusion matrix can be found in Appendix~~\ref{sec:app_results}. %

In addition, we report the ROUGE score~\cite{lin-2004-rouge}, a metric commonly used in summarization evaluation, as a relaxed version of exact match score. The ROUGE scores are omitted in result analysis section, and can be found in Appendix~\ref{sec:app_results}.

\paragraph{Multimodal LLMs.}
We evaluate three flagship proprietary models: GPT-4o~\cite{openai_chatgpt_2024}, Gemini-2.5-Pro~\cite{gemini}, Claude-Sonnet-4~\cite{claude};\footnote{GPT-4o does not internally implement long Chain-of-Though reasoning, while Gemini-2.5-Pro and Claude-Sonnet-4 implement long CoT reasoning (named Thinking Mode) by default. For fair comparison, we disabled Thinking Mode in Claude; Gemini-2.5-Pro does not support full disabling Thinking Mode, we set the thinking token budget to the lowest possible value 128.} and five recent open-source models that can process multiple images, of sizes ranging from 5B to 12B: Phi-4-Multimodal-Instruct (5.6B, \citealp{microsoft2025phi4minitechnicalreportcompact}), Qwen2.5-VL-7B-Instruct~\cite{Qwen2VL}, InternVL-3-8B~\cite{zhu2025internvl3exploringadvancedtraining}, Llama-3.2-11B-Vision-Instruct~\cite{grattafiori2024llama3herdmodels}, and Pixtral-12B~\cite{pixtral}. We use the default inference parameters given by each model in generation.

\begin{table}[t]
    \centering
    \begin{tabular}{lrrrrr}
        \toprule
        \textbf{CLIP} &@1 &@5 &@10 &@20 &@30 \\
        \midrule
        Recall  &2.81   &10.26   &16.70  &25.54  &33.16\\
        NDCG  &24.33	&25.97	&30.98	&39.21	&46.25\\
        Hit &24.33	&53.74	&67.65	&77.54	&82.62\\
        Hit Count &0.24	&1.01	&1.84	&3.18	&4.38\\

        \midrule
        \textbf{BGE} &@1 &@5 &@10 &@20 &@30 \\
        \midrule
        Recall  &3.63	&12.43	&18.81	&28.64	&35.97\\
        NDCG     &26.74	&29.15	&33.4	&41.83	&48.84\\
        Hit    &26.74	&58.56	&68.98	&79.14	&83.96\\
        Hit Count   &0.27	&1.17	&2.04	&3.43	&4.60\\
        \midrule
        \textbf{VLM2Vec} &@1 &@5 &@10 &@20 &@30 \\
        \midrule
        Recall  &2.91	&10.01	&16.03	&26.37	&33.84\\
        NDCG    &24.87	&26.33	&30.09	&39.13	&45.93\\
        Hit    &24.87	&51.87	&62.83	&78.34	&82.35\\
        Hit Count   &0.25	&1.10	&1.91	&3.33	&4.49\\
        \bottomrule
    \end{tabular}
    \caption{Retrieval results using the sub-corpus of 200-300 images for each query. Even within the small, species level corpus, the model struggles with our challenging text-to-image retrieval task.}
    \label{tab:ret_result}
\end{table}



\subsection{Results of Text-to-Image Retrieval}
\label{sec:ret_result}


\begin{table*}[t]
    \centering
    \begin{tabular}{c|ccccc|ccc}
         \toprule
         Model&Phi4-MM&Qwen2.5VL&InternVL3&Pixtral&Llama3.2-V&GPT-4o&Gemini&Claude\\[-0.2em]
         \midrule
         \multicolumn{9}{c}{\textit{Baselines}}\\
         \midrule
    Zero-shot (no image) &35.16	&38.90	&39.17	&41.71	&32.35	&53.74	&60.43	&54.28\\
	GT clue (1 image)  &45.04	&41.79	&43.69	&47.11	&47.81	&59.81	&62.88	&56.79\\
	Non-clue (1 image)  &34.76	&30.08	&29.28	&42.74	&40.37	&14.97	&17.11	&21.39\\
         \midrule
         \multicolumn{9}{c}{\textit{Top-K Retrieval-Augmented Generation}}\\
         \midrule
	k=1    &\cc{g}{39.17}	&\cc{g2}{41.98}	&\cc{r}{37.57}	&\cc{g}{42.11}	&\cc{g}{44.25}	& \cc{r}{24.06}	&\cc{r}{32.22}	&\cc{r}{35.03}\\
	3      &\cc{g}{39.30}	&\cc{g2}{44.39}	&\cc{g}{41.18}	&\cc{r}{39.71}	&\cc{g}{\textbf{46.79}}	&\cc{r}{41.18}	&\cc{r}{48.53}	&\cc{r}{45.32}\\
	5      &\cc{g}{37.70}	&\cc{g2}{46.93}	&\cc{g}{39.97}	&\cc{r}{41.71}	&\cc{g}{43.80}	&\cc{r}{47.86}	&\cc{r}{54.95}	&\cc{r}{50.40}\\
	7      &\cc{g}{\textbf{41.44}}	&\cc{g2}{45.99}	&\cc{g}{\textbf{42.65}}	&\cc{r}{41.44}	&\cc{g}{43.98}	&\cc{r}{51.34}	&\cc{r}{55.35}	&\cc{r}{52.14}\\
	10     &\cc{g}{\textbf{41.44}}	&\cc{g2}{48.26}	&\cc{g}{42.11}	&\cc{g}{42.65}	&\cc{g}{43.42}	&\cc{r}{49.73}	&\cc{r}{57.62}	&\cc{r}{53.88}\\
	15     &\cc{g}{41.04}	&\cc{g2}{49.73}	&\cc{g}{41.04}	&\cc{g}{\textbf{44.39}}   &\cc{g}{44.92}	&\cc{r}{50.80}	&\cc{r}{60.03}	&\cc{g}{55.48}\\
	20     &\cc{g}{41.31}	&\cc{g2}{\textbf{50.53}}	&\cc{g}{41.31}	&-	&-	&\cc{r}{\textbf{52.67}}	&\cc{g}{\textbf{61.50}}	&\cc{g2}{\textbf{57.35}}\\
        \midrule
        \multicolumn{9}{c}{\textit{One-in-K Augmented Generation}}\\
        \midrule
k=3    &\cc{g}{\textbf{41.92}}	&\cc{g2}{46.85}	&\cc{g}{\textbf{41.06}}	&\cc{g}{\textbf{47.11}}	&\cc{g}{\textbf{46.73}}	&\cc{r}{48.95}	&\cc{r}{59.57}	&\cc{r}{48.05}\\
5      &\cc{g}{40.13}	&\cc{g2}{48.04}	&\cc{r}{38.44}	&\cc{g}{46.68}	&\cc{g}{45.19}	&\cc{r}{53.22}	&\cc{g}{60.70}	&\cc{r}{49.86}\\
7      &\cc{g}{40.80}	&\cc{g2}{47.78}	&\cc{g}{39.76}	&\cc{g}{46.68}	&\cc{g}{44.20}	&\cc{g}{55.36}	&\cc{g}{61.53}	&\cc{r}{51.01}\\
10     &\cc{g}{39.08}	&\cc{g2}{47.43}	&\cc{r}{38.50}	&\cc{g}{44.83}	&\cc{g}{41.90}	&\cc{g}{56.25}	&\cc{g}{62.70}	&\cc{r}{51.69}\\
15     &\cc{g}{40.40}	&\cc{g2}{\textbf{49.57}}	&\cc{r}{39.14}	&\cc{g}{43.95}	&\cc{g}{41.12}	&\cc{g}{56.36}	&\cc{g2}{\textbf{64.01}}	&\cc{r}{\textbf{52.67}}\\
20     &\cc{g}{40.83}	&\cc{g2}{48.55}	&\cc{r}{39.09}	&-	&-	&\cc{g}{\textbf{56.50}}	&\cc{g2}{63.47}	&\cc{r}{52.14}\\
        \bottomrule
    \end{tabular}
    \caption{Main experiment results. The coloured cells shows the difference with zero-shot score,  \textcolor{pink}{pink} cells indicate performance under zero-shot baseline, \textcolor{g}{light green} cells indicate performance over zero-shot, but lower than GT, and \textcolor{g2}{green} cells indicate outperforming GT. All models benefit from ground-truth clue image as augmentation.}
    \label{tab:main_results}
\end{table*}

\paragraph{Overall difficulty.}
Successfully retrieving clue images that contain the required visual knowledge is critical to the overall performance of RAG systems in answering the questions. As shown in Table~\ref{tab:ret_result}, across the three retrievers, performance is tightly clustered, with \textsc{BGE} slightly ahead. Even when restricting the search space to 200–300 images per species, retrieval remains challenging: across all retrievers, there is $\sim$ 17-18\% chance that no clue image appears within the top‑30. These results indicate that (i) clue images are genuinely rare and (ii) clue images, if retrieved, are not ranked top by similarity scores. The sub-optimal performance of retrievers poses a further challenge for MLLMs, requiring models to tackle with the task under noisy image context.

\paragraph{Takeaway.}
Substantial performance gap remains to be closed for cross-modal retriever. Performance remains far from saturated even at $k=30$, downstream gains will depend not only on the MLLM’s visual reasoning but also on the retriever exposing more clues, and earlier in the ranking.

\subsection{Results of Visual Retrieval-Augmented Generation}
\label{sec:rag_result}

\paragraph{Does image as augmentation benefit MLLMs?} 

To assess whether image can be leveraged as evidence in the RAG system, we conduct baseline experiments and compare performance under three conditions: 1) \textbf{Zero-shot}: directly prompting MLLMs with text questions without image retrieval; 2) \textbf{GT clue}: augmenting the question with one ground-truth clue image;\footnote{Results reported are the average of 5 runs with different random clue image selected when possible.} 3) \textbf{Non-clue}: augmenting each question with a non-clue image within the same species, which does not contain visual knowledge relevant to the question, as shown in the first section of Table \ref{tab:main_results}.

GT clue setting establishes an upper bound on how images can contribute in RAG. Relative to zero-shot baseline, supplying a single ground-truth clue image raises performance for all models tested. Open-source models receive up to 15 accuracy scores gain, confirming that augmenting relevant visual evidence can substantially compensate for gaps in their pre-trained prior knowledge. The three proprietary models, while having strong zero-shot baseline performance, also improved with augmenting clue image for 2-6 points in accuracy. Non-clue image, conversely, degrades performance of most models, underscoring that the benefit comes from evidence encoded in the clue itself, not from the mere presence of an image.

\textbf{Takeaway: } A single, representative clue image reliably improves performance, confirming visual evidence in our benchmark is informative and influential for RAG.

\paragraph{Does realistic visual RAG benefit MLLMs?}

We have shown that a relevant clue image can serve as good evidence in augmentation. However, retrieval performance for these challenging queries is far from perfect, as shown in the previous section. For RAG system in real implementation where a number of possibly irrelevant images are retrieved, does retrieval still help the final QA performance?

To explore this realistic RAG scenario, we provide the top-$k$ retrieved images,\footnote{For Pixtral and Llama3.2-V, $k=20$ was not evaluated due to hardware limitation.} derived from Section \ref{sec:ret_result}, as augmentation to the MLLMs. 
The retrieved images are sorted in descending order based on their similarity scores. 

Intuitively, more retrieved images would lead to better performance, as probability of retrieving clue image increase with $k$. 
As shown in the second section of Table \ref{tab:main_results}, while Qwen and the three proprietary models follow the expected trend, other open-source models' performance saturates and plateaus at medium $k$ (5 - 10), including more retrieved images offers little benefit and sometimes slightly hinders the performance.

Two intertwined factors plausibly explain the divergence: the open-source models have limited multiple-image processing capacity, as $k$ increase beyond 15, they struggle with effectively extracting related visual information; in the meantime, while larger $k$ brings higher probability of retrieving clue images, the number of negative non-clue images occupy larger portion of the image set. This further dilutes the clue signal and magnifies the impact of their restricted multi-image understanding ability.

\textbf{Takeaway: } Under realistic RAG setting, most of the evaluated models fall noticeably short of the GT clue upper bound, revealing current MLLMs' limitation in \emph{efficiently} utilizing retrieved image context. However, Qwen stands as a powerful exception, possessing robust multimodal RAG ability.

\paragraph{Does a guaranteed clue outweigh the noise of additional non-clue images?}


We notice that top-$k$ RAG underperform the GT clue setting for most models. Such results conflate two error sources: (i) the retriever may fail to return any clue image, or (ii) clue present, but models fail to extract information alongside visually similar non-clue distractors.
To investigate further, we conduct an additional experiment named 1-in-$k$, constructing image sets in which exactly one clue image and $k-1$ non-clue images are provided,\footnote{5 runs for different random images combinations at each $k$.} isolating the impact from imperfect retrieval. The clue image is positioned at the beginning of the $k$ images, emulating perfect retrieval scenario.

Demonstrated in the last section of Table \ref{tab:main_results}, most open-source models except Qwen now lose accuracy as $k$ increases. Proprietary models, in stark contrast, keep improving monotonically with $k$, while GPT-4o and Gemini surpass their top-$k$ RAG performance. Hence, the plateau observed in earlier RAG experiments for open-source models is not merely a retrieval problem: even when the clue is guaranteed, models are still prone to more distractors. Conversely, the unsatisfactory performance of proprietary models at lower $k$ under top-$k$ RAG can be confirmed to be attributable from missing clue image retrieved.

\textbf{Takeaway: } With a guaranteed clue image, proprietary models remain robust to additional non-clue distractors, underscoring their ability to extract visual evidence; whereas open-source models are markedly distractor-sensitive.

\begin{figure}[t]
    \centering
    \includegraphics[width=\linewidth]{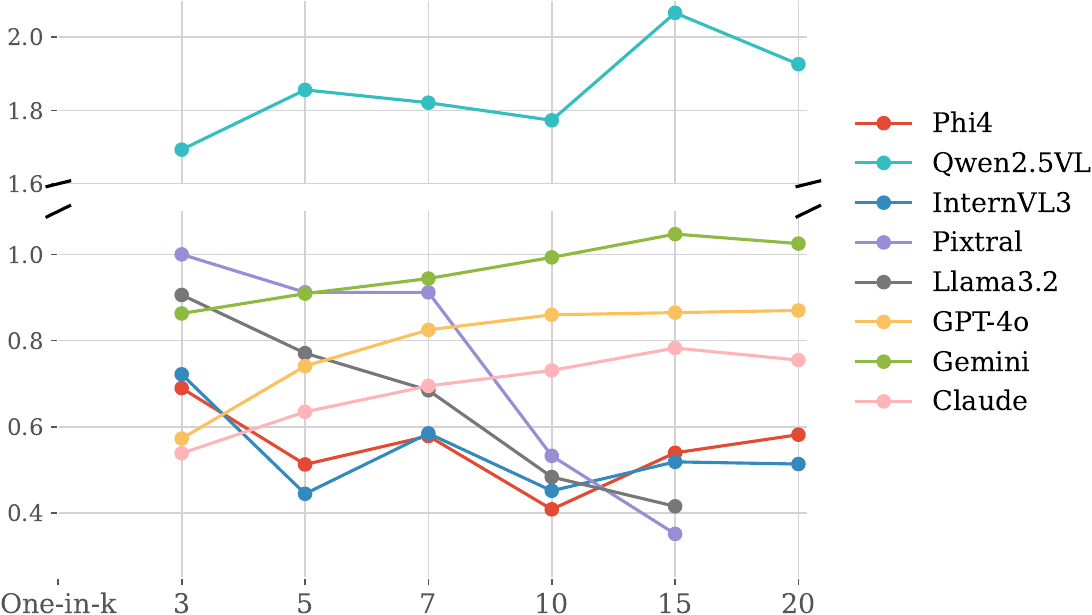}
    \caption{gCUE metrics by models and $k$.}
    \label{fig:gcue}
\end{figure}

\paragraph{How efficiently do MLLMs utilize clue and how robust are they against distractors?} 
With the 1-in-$k$ results, we propose a quantitatively metric named generalized clue utilization efficiency (gCUE). Specifically, we integrate zero-shot and non-clue baselines via $B^{(\lambda)}=\lambda A_Z+(1-\lambda)A_{NC}$, where $\lambda\in[0,1]$, $A_Z$ denotes zero-shot accuracy, $A_{NC}$ denotes non-clue accuracy, $A_{GT}$ denotes GT clue accuracy and $A_k$ denotes 1-in-$k$ accuracy at $k$. The non-clue baseline $A_{NC}$ is introduced as an approximation of the impact on performance from non-clue images. We define:
\[
\mathrm{gCUE}(k;\lambda)=\frac{A_{k}-B^{(\lambda)}}{A_{GT}-B^{(\lambda)}}.
\]
$\mathrm{gCUE}$ therefore quantifies the remaining benefit of the clue image over baseline among $k-1$ distractors, compared with sole clue image scenario. It describes how effectively the model utilizes the visual clue and how it is prone to distractors. $\mathrm{gCUE}=1$ implies model attaining full GT clue benefit, $\mathrm{gCUE}<1$ indicates under-utilization of the clue and vulnerability to distractors, and $\mathrm{gCUE}>1$ suggests model gaining synergy with non-clues. By default we set $\lambda=0.5$.

As shown in Figure~\ref{fig:gcue}, Qwen demonstrates strong synergy and distractor tolerance across $k$. Other open-source models generally display a trend of decline. Their ability to exploit the single clue erodes, suggesting limited capacity for multi-image aggregation and greater vulnerability to a rising distractor load. Whereas proprietary models start with low $\mathrm{gCUE}$ at small $k$, implying that a handful of non-clue images can initially overwhelm the clue despite being positioned at front. Yet they surprisingly recover as more non-clue context being provided. One plausible explanation is that the proprietary models and Qwen implicitly contrast relevant and irrelevant visual features among the noisy image set. As more non-clues are added, the distinctive features of the clue become easier to isolate, eventually lead to confidence in extracting knowledge from the clue image. A detailed causal analysis of this effect is beyond the scope of the present paper and we leave it to future work.

\textbf{Takeaway: } $\mathrm{gCUE}$ metric provides an interpretable metric apart from raw accuracy. Open-source models except Qwen continue to exhibit inverted dynamics comparing to proprietary models on clue utilization efficiency.

\section{Conclusion}

In this paper, we present \dsns, a visual knowledge-intensive VQA benchmark that requires models extracting evidence encoded in images to answer questions. Different from previous works that include images as part of queries and let models retrieve textual knowledge, our \ds is formed with text-only queries and enforces text-to-image retrieval for challenging visual knowledge grounded question answering. We evaluate eight popular MLLMs under various multimodal RAG settings. The results indicate that images can serve as knowledge sources for answering visual knowledge-intensive questions, but the models struggle to effectively extract visual knowledge from noisy retrieved contexts with hard negatives introduced. We further quantify model behavior with a unified efficiency metric $\mathrm{gCUE}$, measuring models' robustness at different numbers of images. As one of the pioneer works in visual-knowledge centric multimodal RAG, our benchmark can help promote the research and development of MLLMs in extracting fine-grained visual knowledge and enhance cross-modal retrieval in a more realistic RAG setting.

\section*{Ethical Statement}
\paragraph{Licensing}
We do not own, nor do we redistribute the images of iNaturalist 2021 dataset. \ds only provides the annotations of queries and answers, as well as image labels on whether it is clue to query, linked to image GUID in iNatrualist 2021 dataset. All images in iNat21 dataset are shared under one of the following Creative Commons licenses, in particular: CC BY 4.0, CC BY-NC 4.0, CC BY-NC-ND 4.0, CC BY-NC-SA 4.0, CC0 1.0, CC BY-ND 4.0, CC BY-SA 4.0. Any usage of the images is subject to iNaturalist and iNaturalist 2021 dataset's term of use. We strictly adhere to the intended non-commercial research use granted by the iNaturalist community and iNaturalist 2021 dataset, and we share our annotation in \ds under CC BY-NC 4.0 license. The authors assume no responsibility for any non-compliant usage or legal/ethical issues related to the original iNaturalist 2021 dataset.

\paragraph{Potential Risks}
The annotation in this dataset, i.e. the queries, answers and image labels, are intended for research purposes only. Our dataset’s queries, answers, and image labels are provided by volunteer annotators who are not professional taxonomists or biologists. We make every effort to ensure correctness, such as proactively consulting publicly available expertised resources of biology and taxonomy over internet, and annotators are undergraduate students who possess sufficient English proficiency to utilize such publicly available resources for fact validation. Though, we cannot guarantee the absence of factual errors. We make no guarantees regarding completeness or correctness for real-world decision-making.

Our annotation does not contain any personally identifiable information, nor does it reference private data, locations, or individuals. Our annotation does not reveal exact geographical coordinates or sensitive ecological information that might pose risks to endangered species or protected habitats. For other potential risks regarding images in iNaturalist 2021 dataset, please refer to the original paper and iNaturalist website (www.inaturalist.org).

\bibliography{bibl}

\begin{thebibliography}{35}
\providecommand{\natexlab}[1]{#1}

\bibitem[{Anthropic(2025)}]{claude}
Anthropic. 2025.
\newblock Introducing Claude 4.
\newblock \url{https://www.anthropic.com/news/claude-4}.
\newblock Accessed: 2025-07-25.

\bibitem[{Bai et~al.(2025)Bai, Chen, Liu, Wang, Ge, Song, Dang, Wang, Wang, Tang, Zhong, Zhu, Yang, Li, Wan, Wang, Ding, Fu, Xu, Ye, Zhang, Xie, Cheng, Zhang, Yang, Xu, and Lin}]{Qwen2VL}
Bai, S.; Chen, K.; Liu, X.; Wang, J.; Ge, W.; Song, S.; Dang, K.; Wang, P.; Wang, S.; Tang, J.; Zhong, H.; Zhu, Y.; Yang, M.; Li, Z.; Wan, J.; Wang, P.; Ding, W.; Fu, Z.; Xu, Y.; Ye, J.; Zhang, X.; Xie, T.; Cheng, Z.; Zhang, H.; Yang, Z.; Xu, H.; and Lin, J. 2025.
\newblock Qwen2.5-VL Technical Report.
\newblock arXiv:2502.13923.

\bibitem[{Chang et~al.(2022)Chang, Narang, Suzuki, Cao, Gao, and Bisk}]{Chang_2022_CVPR}
Chang, Y.; Narang, M.; Suzuki, H.; Cao, G.; Gao, J.; and Bisk, Y. 2022.
\newblock WebQA: Multihop and Multimodal QA.
\newblock In \emph{Proceedings of the IEEE/CVF Conference on Computer Vision and Pattern Recognition (CVPR)}, 16495--16504.

\bibitem[{Chen et~al.(2023)Chen, Hu, Luan, Sun, Changpinyo, Ritter, and Chang}]{chen-etal-2023-pre-trained}
Chen, Y.; Hu, H.; Luan, Y.; Sun, H.; Changpinyo, S.; Ritter, A.; and Chang, M.-W. 2023.
\newblock Can Pre-trained Vision and Language Models Answer Visual Information-Seeking Questions?
\newblock In Bouamor, H.; Pino, J.; and Bali, K., eds., \emph{Proceedings of the 2023 Conference on Empirical Methods in Natural Language Processing}, 14948--14968. Singapore: Association for Computational Linguistics.

\bibitem[{Douze et~al.(2024)Douze, Guzhva, Deng, Johnson, Szilvasy, Mazaré, Lomeli, Hosseini, and Jégou}]{douze2024faisslibrary}
Douze, M.; Guzhva, A.; Deng, C.; Johnson, J.; Szilvasy, G.; Mazaré, P.-E.; Lomeli, M.; Hosseini, L.; and Jégou, H. 2024.
\newblock The Faiss library.
\newblock arXiv:2401.08281.

\bibitem[{Google(2025)}]{gemini}
Google. 2025.
\newblock Gemini-2.5-Pro.
\newblock \url{https://deepmind.google/models/gemini/pro/}.
\newblock Accessed: 2025-07-25.

\bibitem[{Hu et~al.(2023)Hu, Luan, Chen, Khandelwal, Joshi, Lee, Toutanova, and Chang}]{Hu_2023_ICCV}
Hu, H.; Luan, Y.; Chen, Y.; Khandelwal, U.; Joshi, M.; Lee, K.; Toutanova, K.; and Chang, M.-W. 2023.
\newblock Open-domain Visual Entity Recognition: Towards Recognizing Millions of Wikipedia Entities.
\newblock In \emph{Proceedings of the IEEE/CVF International Conference on Computer Vision (ICCV)}, 12065--12075.

\bibitem[{Hu et~al.(2025)Hu, Gu, Dou, Fayyaz, Lu, Chang, and Peng}]{hu2024mragbenchvisioncentricevaluationretrievalaugmented}
Hu, W.; Gu, J.-C.; Dou, Z.-Y.; Fayyaz, M.; Lu, P.; Chang, K.-W.; and Peng, N. 2025.
\newblock {MRAG}-Bench: Vision-Centric Evaluation for Retrieval-Augmented Multimodal Models.
\newblock In \emph{The Thirteenth International Conference on Learning Representations}.

\bibitem[{Huang et~al.(2024)Huang, Qu, Bu, Zhou, Liu, Yang, Xu, and Zhao}]{huang2024empiricalstudyllmasajudgellm}
Huang, H.; Qu, Y.; Bu, X.; Zhou, H.; Liu, J.; Yang, M.; Xu, B.; and Zhao, T. 2024.
\newblock An Empirical Study of LLM-as-a-Judge for LLM Evaluation: Fine-tuned Judge Model is not a General Substitute for GPT-4.
\newblock arXiv:2403.02839.

\bibitem[{Izacard and Grave(2021)}]{izacard-grave-2021-leveraging}
Izacard, G.; and Grave, E. 2021.
\newblock Leveraging Passage Retrieval with Generative Models for Open Domain Question Answering.
\newblock In Merlo, P.; Tiedemann, J.; and Tsarfaty, R., eds., \emph{Proceedings of the 16th Conference of the European Chapter of the Association for Computational Linguistics: Main Volume}, 874--880. Online: Association for Computational Linguistics.

\bibitem[{Jiang et~al.(2025)Jiang, Meng, Yang, Yavuz, Zhou, and Chen}]{jiang2025vlmvec}
Jiang, Z.; Meng, R.; Yang, X.; Yavuz, S.; Zhou, Y.; and Chen, W. 2025.
\newblock {VLM}2Vec: Training Vision-Language Models for Massive Multimodal Embedding Tasks.
\newblock In \emph{The Thirteenth International Conference on Learning Representations}.

\bibitem[{Kamalloo et~al.(2023)Kamalloo, Dziri, Clarke, and Rafiei}]{kamalloo-etal-2023-evaluating}
Kamalloo, E.; Dziri, N.; Clarke, C.; and Rafiei, D. 2023.
\newblock Evaluating Open-Domain Question Answering in the Era of Large Language Models.
\newblock In Rogers, A.; Boyd-Graber, J.; and Okazaki, N., eds., \emph{Proceedings of the 61st Annual Meeting of the Association for Computational Linguistics (Volume 1: Long Papers)}, 5591--5606. Toronto, Canada: Association for Computational Linguistics.

\bibitem[{Karpukhin et~al.(2020)Karpukhin, Oguz, Min, Lewis, Wu, Edunov, Chen, and Yih}]{karpukhin-etal-2020-dense}
Karpukhin, V.; Oguz, B.; Min, S.; Lewis, P.; Wu, L.; Edunov, S.; Chen, D.; and Yih, W.-t. 2020.
\newblock Dense Passage Retrieval for Open-Domain Question Answering.
\newblock In Webber, B.; Cohn, T.; He, Y.; and Liu, Y., eds., \emph{Proceedings of the 2020 Conference on Empirical Methods in Natural Language Processing (EMNLP)}, 6769--6781. Online: Association for Computational Linguistics.

\bibitem[{Lerner et~al.(2022)Lerner, Ferret, Guinaudeau, Le~Borgne, Besan\c{c}on, Moreno, and Lov\'{o}n~Melgarejo}]{10.1145/3477495.3531753}
Lerner, P.; Ferret, O.; Guinaudeau, C.; Le~Borgne, H.; Besan\c{c}on, R.; Moreno, J.~G.; and Lov\'{o}n~Melgarejo, J. 2022.
\newblock ViQuAE, a Dataset for Knowledge-based Visual Question Answering about Named Entities.
\newblock In \emph{Proceedings of the 45th International ACM SIGIR Conference on Research and Development in Information Retrieval}, SIGIR '22, 3108–3120. New York, NY, USA: Association for Computing Machinery.
\newblock ISBN 9781450387323.

\bibitem[{Lewis et~al.(2020)Lewis, Perez, Piktus, Petroni, Karpukhin, Goyal, K\"{u}ttler, Lewis, Yih, Rockt\"{a}schel, Riedel, and Kiela}]{NEURIPS2020_6b493230}
Lewis, P.; Perez, E.; Piktus, A.; Petroni, F.; Karpukhin, V.; Goyal, N.; K\"{u}ttler, H.; Lewis, M.; Yih, W.-t.; Rockt\"{a}schel, T.; Riedel, S.; and Kiela, D. 2020.
\newblock Retrieval-Augmented Generation for Knowledge-Intensive NLP Tasks.
\newblock In Larochelle, H.; Ranzato, M.; Hadsell, R.; Balcan, M.; and Lin, H., eds., \emph{Advances in Neural Information Processing Systems}, volume~33, 9459--9474. Curran Associates, Inc.

\bibitem[{Lin(2004)}]{lin-2004-rouge}
Lin, C.-Y. 2004.
\newblock {ROUGE}: A Package for Automatic Evaluation of Summaries.
\newblock In \emph{Text Summarization Branches Out}, 74--81. Barcelona, Spain: Association for Computational Linguistics.

\bibitem[{Lin et~al.(2023)Lin, Asai, Li, Oguz, Lin, Mehdad, Yih, and Chen}]{lin-etal-2023-train}
Lin, S.-C.; Asai, A.; Li, M.; Oguz, B.; Lin, J.; Mehdad, Y.; Yih, W.-t.; and Chen, X. 2023.
\newblock How to Train Your Dragon: Diverse Augmentation Towards Generalizable Dense Retrieval.
\newblock In Bouamor, H.; Pino, J.; and Bali, K., eds., \emph{Findings of the Association for Computational Linguistics: EMNLP 2023}, 6385--6400. Singapore: Association for Computational Linguistics.

\bibitem[{Liu et~al.(2025)Liu, Zhu, Zhou, Zhang, Yi, Yan, Gu, Yu, and Sun}]{liu2025benchmarkingretrievalaugmentedgenerationmultimodal}
Liu, Z.; Zhu, X.; Zhou, T.; Zhang, X.; Yi, X.; Yan, Y.; Gu, Y.; Yu, G.; and Sun, M. 2025.
\newblock Benchmarking Retrieval-Augmented Generation in Multi-Modal Contexts.
\newblock arXiv:2502.17297.

\bibitem[{Marino et~al.(2019)Marino, Rastegari, Farhadi, and Mottaghi}]{Marino_2019_CVPR}
Marino, K.; Rastegari, M.; Farhadi, A.; and Mottaghi, R. 2019.
\newblock OK-VQA: A Visual Question Answering Benchmark Requiring External Knowledge.
\newblock In \emph{Proceedings of the IEEE/CVF Conference on Computer Vision and Pattern Recognition (CVPR)}.

\bibitem[{Mensink et~al.(2023)Mensink, Uijlings, Castrejon, Goel, Cadar, Zhou, Sha, Araujo, and Ferrari}]{Mensink_2023_ICCV}
Mensink, T.; Uijlings, J.; Castrejon, L.; Goel, A.; Cadar, F.; Zhou, H.; Sha, F.; Araujo, A.; and Ferrari, V. 2023.
\newblock Encyclopedic VQA: Visual Questions About Detailed Properties of Fine-Grained Categories.
\newblock In \emph{Proceedings of the IEEE/CVF International Conference on Computer Vision (ICCV)}, 3113--3124.

\bibitem[{{Meta Llama Team}(2024)}]{grattafiori2024llama3herdmodels}
{Meta Llama Team}. 2024.
\newblock The Llama 3 Herd of Models.
\newblock arXiv:2407.21783.

\bibitem[{Microsoft(2025)}]{microsoft2025phi4minitechnicalreportcompact}
Microsoft. 2025.
\newblock Phi-4-Mini Technical Report: Compact yet Powerful Multimodal Language Models via Mixture-of-LoRAs.
\newblock arXiv:2503.01743.

\bibitem[{Mishra et~al.(2022)Mishra, Suryavardan, Bhaskar, Chopra, Reganti, Patwa, Das, Chakraborty, Sheth, Ekbal et~al.}]{mishra2022factify}
Mishra, S.; Suryavardan, S.; Bhaskar, A.; Chopra, P.; Reganti, A.~N.; Patwa, P.; Das, A.; Chakraborty, T.; Sheth, A.~P.; Ekbal, A.; et~al. 2022.
\newblock FACTIFY: A Multi-Modal Fact Verification Dataset.
\newblock In \emph{De-Factify: Workshop on Multimodal Fact Checking and Hate Speech Detection, co-located with AAAI 2022.}

\bibitem[{{Mistral AI}(2024)}]{pixtral}
{Mistral AI}. 2024.
\newblock Announcing Pixtral 12B.
\newblock \url{https://mistral.ai/news/pixtral-12b/}.
\newblock Accessed: 2025-07-25.

\bibitem[{OpenAI(2025)}]{openai_chatgpt_2024}
OpenAI. 2025.
\newblock ChatGPT.
\newblock \url{https://openai.com/chatgpt}.
\newblock Accessed: 2025-07-25.

\bibitem[{Radford et~al.(2021)Radford, Kim, Hallacy, Ramesh, Goh, Agarwal, Sastry, Askell, Mishkin, Clark, Krueger, and Sutskever}]{pmlr-v139-radford21a}
Radford, A.; Kim, J.~W.; Hallacy, C.; Ramesh, A.; Goh, G.; Agarwal, S.; Sastry, G.; Askell, A.; Mishkin, P.; Clark, J.; Krueger, G.; and Sutskever, I. 2021.
\newblock Learning Transferable Visual Models From Natural Language Supervision.
\newblock In Meila, M.; and Zhang, T., eds., \emph{Proceedings of the 38th International Conference on Machine Learning}, volume 139 of \emph{Proceedings of Machine Learning Research}, 8748--8763. PMLR.

\bibitem[{Schwenk et~al.(2022)Schwenk, Khandelwal, Clark, Marino, and Mottaghi}]{10.1007/978-3-031-20074-8_9}
Schwenk, D.; Khandelwal, A.; Clark, C.; Marino, K.; and Mottaghi, R. 2022.
\newblock A-OKVQA: A Benchmark for Visual Question Answering Using World Knowledge.
\newblock In \emph{Computer Vision – ECCV 2022: 17th European Conference, Tel Aviv, Israel, October 23–27, 2022, Proceedings, Part VIII}, 146–162. Berlin, Heidelberg: Springer-Verlag.
\newblock ISBN 978-3-031-20073-1.

\bibitem[{Tanaka et~al.(2023)Tanaka, Nishida, Nishida, Hasegawa, Saito, and Saito}]{10.1609/aaai.v37i11.26598}
Tanaka, R.; Nishida, K.; Nishida, K.; Hasegawa, T.; Saito, I.; and Saito, K. 2023.
\newblock SlideVQA: a dataset for document visual question answering on multiple images.
\newblock In \emph{Proceedings of the Thirty-Seventh AAAI Conference on Artificial Intelligence and Thirty-Fifth Conference on Innovative Applications of Artificial Intelligence and Thirteenth Symposium on Educational Advances in Artificial Intelligence}, AAAI'23/IAAI'23/EAAI'23. AAAI Press.
\newblock ISBN 978-1-57735-880-0.

\bibitem[{Van~Horn et~al.(2021)Van~Horn, Cole, Beery, Wilber, Belongie, and Mac~Aodha}]{Van_Horn_2021_CVPR}
Van~Horn, G.; Cole, E.; Beery, S.; Wilber, K.; Belongie, S.; and Mac~Aodha, O. 2021.
\newblock Benchmarking Representation Learning for Natural World Image Collections.
\newblock In \emph{Proceedings of the IEEE/CVF Conference on Computer Vision and Pattern Recognition (CVPR)}, 12884--12893.

\bibitem[{Van~Landeghem et~al.(2023)Van~Landeghem, Tito, Borchmann, Pietruszka, Joziak, Powalski, Jurkiewicz, Coustaty, Anckaert, Valveny, Blaschko, Moens, and Stanislawek}]{Van_Landeghem_2023_ICCV}
Van~Landeghem, J.; Tito, R.; Borchmann, {\L}.; Pietruszka, M.; Joziak, P.; Powalski, R.; Jurkiewicz, D.; Coustaty, M.; Anckaert, B.; Valveny, E.; Blaschko, M.; Moens, S.; and Stanislawek, T. 2023.
\newblock Document Understanding Dataset and Evaluation (DUDE).
\newblock In \emph{Proceedings of the IEEE/CVF International Conference on Computer Vision (ICCV)}, 19528--19540.

\bibitem[{Vendrow et~al.(2024)Vendrow, Pantazis, Shepard, Brostow, Jones, Aodha, Beery, and Horn}]{vendrow2024inquire}
Vendrow, E.; Pantazis, O.; Shepard, A.; Brostow, G.; Jones, K.~E.; Aodha, O.~M.; Beery, S.; and Horn, G.~V. 2024.
\newblock {INQUIRE}: A Natural World Text-to-Image Retrieval Benchmark.
\newblock In \emph{The Thirty-eight Conference on Neural Information Processing Systems Datasets and Benchmarks Track}.

\bibitem[{Yan and Xie(2024)}]{yan-xie-2024-echosight}
Yan, Y.; and Xie, W. 2024.
\newblock {E}cho{S}ight: Advancing Visual-Language Models with {W}iki Knowledge.
\newblock In Al-Onaizan, Y.; Bansal, M.; and Chen, Y.-N., eds., \emph{Findings of the Association for Computational Linguistics: EMNLP 2024}, 1538--1551. Miami, Florida, USA: Association for Computational Linguistics.

\bibitem[{Zheng et~al.(2023)Zheng, Chiang, Sheng, Zhuang, Wu, Zhuang, Lin, Li, Li, Xing, Zhang, Gonzalez, and Stoica}]{zheng2023judging}
Zheng, L.; Chiang, W.-L.; Sheng, Y.; Zhuang, S.; Wu, Z.; Zhuang, Y.; Lin, Z.; Li, Z.; Li, D.; Xing, E.; Zhang, H.; Gonzalez, J.~E.; and Stoica, I. 2023.
\newblock Judging {LLM}-as-a-Judge with {MT}-Bench and Chatbot Arena.
\newblock In \emph{Thirty-seventh Conference on Neural Information Processing Systems Datasets and Benchmarks Track}.

\bibitem[{Zhou et~al.(2024)Zhou, Liu, Liu, Xiao, Wang, Zhao, Zhang, Lian, and Xiong}]{zhou2024megapairs}
Zhou, J.; Liu, Z.; Liu, Z.; Xiao, S.; Wang, Y.; Zhao, B.; Zhang, C.~J.; Lian, D.; and Xiong, Y. 2024.
\newblock MegaPairs: Massive Data Synthesis For Universal Multimodal Retrieval.
\newblock \emph{arXiv preprint arXiv:2412.14475}.

\bibitem[{Zhu et~al.(2025)Zhu, Wang, Chen, Liu, Ye, Gu, Tian, Duan, Su, Shao, Gao, Cui, Wang, Cao, Liu, Wei, Zhang, Wang, Xu, Li, Wang, Deng, Li, He, Jiang, Luo, Wang, He, Shi, Zhang, Shao, He, Xiong, Qu, Sun, Jiao, Lv, Wu, Zhang, Deng, Ge, Chen, Wang, Dou, Lu, Zhu, Lu, Lin, Qiao, Dai, and Wang}]{zhu2025internvl3exploringadvancedtraining}
Zhu, J.; Wang, W.; Chen, Z.; Liu, Z.; Ye, S.; Gu, L.; Tian, H.; Duan, Y.; Su, W.; Shao, J.; Gao, Z.; Cui, E.; Wang, X.; Cao, Y.; Liu, Y.; Wei, X.; Zhang, H.; Wang, H.; Xu, W.; Li, H.; Wang, J.; Deng, N.; Li, S.; He, Y.; Jiang, T.; Luo, J.; Wang, Y.; He, C.; Shi, B.; Zhang, X.; Shao, W.; He, J.; Xiong, Y.; Qu, W.; Sun, P.; Jiao, P.; Lv, H.; Wu, L.; Zhang, K.; Deng, H.; Ge, J.; Chen, K.; Wang, L.; Dou, M.; Lu, L.; Zhu, X.; Lu, T.; Lin, D.; Qiao, Y.; Dai, J.; and Wang, W. 2025.
\newblock InternVL3: Exploring Advanced Training and Test-Time Recipes for Open-Source Multimodal Models.
\newblock arXiv:2504.10479.

\end{thebibliography}

\appendix
\section{Extended Comparison with Related Works}
\label{sec:app_relation}

Knowledge intensive VQA benchmarks focusing textual knowledge, including but not limited to: ViQuAE~\citep{10.1145/3477495.3531753}, InfoSeek~\citep{chen-etal-2023-pre-trained} and Encyclopedic-VQA~\citep{Mensink_2023_ICCV}, are discussed in related works. These benchmarks contain questions paired with images of certain entities, and knowledge for such entities can be found in an external textual knowledge base (e.g., Wikipedia). The questions are deliberately constructed to omit the entity names (e.g., ``How many feet tall does the plant grow to?'' instead of ``How many feet tall does Acacia paradoxa grow to?'').

Typically, given a question and its associated image, a model is expected to: 1) identify the entity depicted in the image, 2) retrieve relevant documents or passages from a knowledge base that pertain to both the question and the entity, and 3) generate an answer based on the question, image, and retrieved documents. Step 1) and 2) is usually carried out by leveraging vision-language pretrained encoders, such as CLIP~\citep{pmlr-v139-radford21a}, to facilitate cross-modal (image-to-text) retrieval.

While the above mentioned benchmarks are all claimed to be knowledge-intensive VQA benchmarks, it is noteworthy that the \textit{knowledge} to be retrieved is textual knowledge. The image is only served as an anchor of the entity. While cross-modal retrieval was commonly used, the knowledge retrieval process can actually be achieved in uni-modal manner. \citet{yan-xie-2024-echosight} achieved ``SoTA'' performance on InfoSeek and E-VQA employing image-to-image retrieval within a knowledge corpus comprising Wikipedia pages and their associated images. Entity recognition is performed by identifying the most visually similar Wikipedia image to the query image, which automatically enables retrieval of the related knowledge document (i.e., the Wikipedia page containing that image). Answers are then generated using a \textbf{text-only} LLM, augmented with the relevant paragraph from the Wikipedia page, which is extracted through fine-grained reranking.

WebQA dataset~\citep{Chang_2022_CVPR} included questions that require retrieving images and utilizing visual knowledge to generate answers. However, its retrieval process is heavily reliant on image captions, as the images themselves typically lack indicators identifying them as clue images for answering a question. For instance, considering the question: ``\textit{Are the land dinosaurs guarded by rail in both the Display Museum of Natural History in University of Michigan and the Museo Jurassic de Asturias?}'' Without captions, the model would be unable to discern which dinosaur fossil image corresponds to which museum.
Table \ref{tab:ret_webqa} shows the retrieval scores in our implementation of retrieval experiments on WebQA benchmark for queries having images as knowledge source, evaluated with clip-vit-large-patch14-336. While the queries are text-only, aligning with our setting, the text-to-image retrieval scores significantly lag behind text-to-caption retrieval, especially for lower $k$.

\begin{table}[t]
    \centering
    \begin{tabular}{l|lllll}
    \toprule
        Recall  & @1  &@5 &@10    &@20    &@50 \\
        \midrule
        T2I &15.17	&37.08	&47.63	&58.43	&72.32\\
        T2T &43.25	&68.74	&76.86	&83.79	&89.92\\

    \bottomrule
    \end{tabular}
    \caption{Retrieval scores for WebQA dataset. T2I: Query-to-Image retrieval. T2T: Query-to-Caption retrieval. T2T retrieval significantly outperform T2I, indicating crucial identifier for images being the clue reside in caption.}
    \label{tab:ret_webqa}
\end{table}

In addition to MRAG-Bench introduced in the main paper, another recent work M$^2$RAG~\cite{liu2025benchmarkingretrievalaugmentedgenerationmultimodal} gathers data samples from existing benchmarks, namely WebQA introduced in previous paragraph and Factify~\cite{mishra2022factify}, a multimodal fact verification dataset. The multi-modal QA subtask introduced partially overlap with ours, while the queries are repackaged instances drawn from the original WebQA dataset.

The iNaturalist team released 2024 version of their natural image dataset, iNat24~\citep{vendrow2024inquire}, coming with a text-to-image retrieval benchmark named INQUIRE, containing 250 queries. Though, the queries in INQUIRE are designated for image retrieval only instead of QA style questions that have explicit answers. As iNat24 and iNat21 share the same taxonomy classes of organisms and contain newly uploaded images not included in iNat21, we do consider adding the new images in iNat24 to our benchmark. We conduct preliminary experiment using iNat24 images as retrieval corpus, evaluated with Qwen2.5VL, as shown in Table~\ref{tab:inat24}. The overall performance remains at comparable level.

\begin{table}[t]
    \centering
    \begin{tabular}{l|rr}
    \toprule
        Top-$k$ RAG& iNat21& iNat24\\
         \midrule
        k=1&    41.98&  42.72\\
        k=3&    44.39&  42.18\\
        k=5&    46.93&  45.96\\
        k=7&    45.99&  46.23\\
        k=10&   48.26&  45.28\\
        k=15&   49.73&  48.47\\
        k=20&   50.53&  48.79\\
    \bottomrule
    \end{tabular}
    \caption{Top-$k$ RAG performance using iNat24 images as knowledge corpus, evaluated with Qwen2.5VL.}
    \label{tab:inat24}
\end{table}

We also note a related line of research~\citep{10.1609/aaai.v37i11.26598,Van_Landeghem_2023_ICCV} on extracting knowledge from document images (e.g., PDF files, Powerpoint slides), conventionally referred to as OCR-based VQA. Although these tasks also require models to retrieve relevant page and interpret images to answer textual queries, the underlying knowledge is primarily textual—obtained via OCR. This focus diverges considerably from our goal of visual feature extraction, where visual cues are inherently more sparse in the wild.

\section{Additional Ablation Studies}

\subsection{Can the answers for \ds be found in Wikipedia?}
We construct a small textual knowledge base from Wikipedia articles for the 10k species in iNat21 dataset. This minimum knowledge base comprising approximately 97,000 chunks, each consisting of $\sim$150 tokens. 

Among the 10000 taxonomy species in iNat21 dataset, we matched 9614 Wikipedia pages. iNat21 provides the scientific name as well as one common name of the species. We conduct multiple rounds of matching: initially, Wikipedia articles are matched based on species scientific names with Wikipedia titles; for unmatched species, match their common names with article titles; further unmatched species will go through matching within article content. All articles containing the scientific name are collected, sorting according to the position of the scientific name's first occurrence within the article. The article with earliest occurrence is selected as matched page. Due to the fact that taxonomy study frequently change the attribution of a species, the scientific name in iNat21 dataset can be deprecated and no matched page can be found for them.

We then applied basic text chunking to these articles, splitting them into chunks of approximately 150 tokens. Each chunk is bounded within a single section of the original article to preserve contextual integrity. Sections such as ``Reference'', ``See also'', ``External links'', etc., which lack informative content were excluded. The chunks are cut upon period mark, so that they do not end or start with incomplete sentence. Two or more adjacent sections having total length less than 150 tokens are merged, with the section titles kept.

For text retrieval, we use a pre-trained dense text retriever, DRAGON Plus~\cite{lin-etal-2023-train}. Given that the ground-truth labels for relevant text chunks are not available, we only assess whether the retrieved text chunks correspond to the correct Wikipedia page, and the retriever achieves Hit@10 of 97\%. Next, we present the QA performance across open-sourced models augmented with top-10 retrieved textual knowledge chunks. As shown in Table \ref{tab:ablation}, augmenting with textual knowledge from Wikipedia result in performance drop for both models tested. Instead, in numerous cases, models fail to locate relevant information within the augmented text and output ``I don't know''. This behaviour in turn leads to scores lower than the zero-shot setting. 

These experiments provide strong evidence that the knowledge required to answer the questions related to visual features in our benchmark is unlikely to be textually documented in Wikipedia.

\begin{table}[t]
    \centering
    \begin{tabular}{r|ccc}
    \toprule
         &  Qwen2.5VL&    GPT-4o\\
         \midrule
        Zero-shot& 40.23&    56.58\\
        RAG-Best&  51.88(@20)  &56.58(@7)\\
        \midrule
        Text RAG&   25.53   &31.68 \\
        \midrule
        1GT&41.73	&59.89	\\
        2GT&48.12	&53.38	\\
        3GT&54.51	&54.89	\\
        4GT&53.20	&56.77	\\
        5GT&52.44	&60.15	\\
        \midrule
        1-in-10&45.90	&56.94	\\
        2-in-10&47.20	&58.08	\\
        3-in-10&51.13	&61.09	\\
        4-in-10&52.19	&59.21	\\
        5-in-10&52.07	&61.47	\\
    \bottomrule
    \end{tabular}
    \caption{Ablation experiment: results when augmenting textual knowledge and impact from number of clue images. 1GT - 5GT: input 1-5 ground truth clue images only ($m$GT setting); 1-in-10: 1 clue with 9 non-clues, 2 clues with 8 non-clues, etc.}
    \label{tab:ablation}
\end{table}

\subsection{Does including more clue images consistently lead to better performance?}
We observed that in most models, top-$k$ RAG scores at higher $k$ values excel their 1-in-$k$ score.  A likely explanation is that the expected number of clue images retrieved in top-$k$ is substantially higher than when only a single clue image is guaranteed. We in turn ablate with effect of multiple clue images. We compare two regimes on the subset of queries with sufficient clues: 
$m$GT ($m$ ground-truth clue images) and $m$-in-10 ($m$ clues mixed with $10{-}m$ non-clues from the same species), reported in Table \ref{tab:ablation}. Aligned with main experiments, mixing non-clues is beneficial at low $m$: non-clue context appears to regularize both models' multi-image aggregation ability, improving robustness when only a few clues are present.  For Qwen, the $m$GT results forms a bell-shaped curve, where 3 clue images gives the best performance, while $m$-in-10 gives overall increasing curve. This suggests Qwen benefits most from a small set of clean, diverse clue views, and is sensitive to additional non-clue context. GPT-4o, conversely, gives a U-shaped curve for $m$GT. Interestingly, small number of clue images corrupts the model's correct answer when only one clue is present, and mixing them with non-clues brings the performance close to optimal.

\begin{table}[t]
    \centering
    \begin{tabular}{l|rr}
    \toprule
        & Same Species& Different Species\\
         \midrule
        k=3&    46.85&  46.15\\
        k=5&    48.04&  47.41\\
        k=7&    47.78&  46.97\\
        k=10&   47.43&  47.55\\
        k=15&   49.57&  47.72\\
        k=20&   48.55&  47.76\\
    \bottomrule
    \end{tabular}
    \caption{Impact from using highly irrelevant distractor in 1-in-k setting, evaluated with Qwen2.5VL.} 
    \label{tab:diff_distractor}
\end{table}
\subsection{Hard negatives can be helpful; how about easy negatives?}
We observed that for Qwen2.5VL and some proprietary models, the performance of mixing non-clues with a clue image can sometimes surpass the one clue image upper-bound. As an ablation to the experiment findings, we now mix one clue image with $k-1$ highly irrelevant images from subsets that are not the query species. The results in Table \ref{tab:diff_distractor} indicate that benefit from hard negatives overall surpass that from easy negatives, especially at higher $k$, partially confirming our hypothesis in last experiment of Section 5.3. As an example, for question ``What color are the hindwings of Papaipema inquaesita?'', hard negatives displays the moth not showing the hindwings, while easy negatives are images of plants, mammals, etc. Without the normal image showing only forewings, model could be confused even when presented the hindwings.

\subsection{Does visual clue reduce ``I don't know'' responses?}
During qualitative inspection we observed that, at small $k$, proprietary models often answer with phrases such as ``I cannot decide'' or ``Not enough information.'' We conventionally name such outputs as ``IDK'' responses. A natural hypothesis is that adding clue images should increase model confidence and therefore lower the IDK rate. Table \ref{tab:idk_rate} confirms this for the proprietary models and for Qwen: as $k$ grows, their IDK rate falls, mirroring the rise in accuracy. By contrast, Phi, Pixtral, and Llama produce few IDK responses even when their accuracy fluctuates, suggesting that these models tend to commit to an answer rather than abstain. This tendency also explains the seemingly counter-intuitive result that Llama’s accuracy in the Non-clue setting exceeds its zero-shot baseline. Although the added non-clue image lacks the required visual evidence, it could provide the model with organism-specific context, prompting it to draw on parametric knowledge to answer.

\begin{table*}[t]
    \centering
    \begin{tabular}{c|ccccc|ccc}
         \toprule
         Model  &Phi4-MM	&Qwen2.5VL	&InternVL3	&Pixtral	&Llama3.2-V	&GPT-4o	&Gemini	&Claude\\
         \midrule
         \multicolumn{9}{c}{\textit{Baselines}}\\
         \midrule
     Zero-shot (no image)   &0.27	&0.00	&1.60	&0.53	&11.76	&0.00	&0.53	&3.74\\
    GT clue (1 image) &0.59	&9.89	&6.52	&0.48	&2.84	&4.60	&2.41	&5.83\\
    Non-clue (1 image)	&2.14	&27.27	&31.28	&1.87	&4.81	&76.74	&71.12	&67.11\\
         \midrule
         \multicolumn{9}{c}{\textit{Top-K Retrieval-Augmented Generation}}\\
         \midrule
k=1&0.80	&10.43	&13.10	&0.80	&2.67	&58.02	&47.86	&41.18\\
3&0.80	&8.29	&8.29	&0.00	&1.07	&27.54	&24.60	&20.86\\
5&0.53	&3.74	&8.82	&0.53	&1.34	&18.45	&16.58	&12.57\\
7&0.80	&3.74	&7.22	&1.07	&0.53	&15.24	&14.17	&11.23\\
10&0.27	&2.94	&6.42	&0.27	&1.60	&15.24	&11.50	&8.82\\
15&2.14	&2.94	&5.08	&0.27	&1.34	&13.64	&9.36	&7.22\\
20&1.87	&2.14	&3.21	&-	&-	&10.70	&6.15	&8.02\\
        \midrule
        \multicolumn{9}{c}{\textit{1-in-K Augmented Generation}}\\
        \midrule
k=3&1.34	&6.42	&10.64	&0.32	&1.02	&21.76	&8.18	&16.36\\
5&1.02	&3.21	&9.25	&0.27	&0.85	&15.08	&5.77	&13.10\\
7&1.07	&2.94	&10.64	&0.37	&1.34	&12.41	&4.49	&10.43\\
10&1.18	&2.40	&10.32	&0.22	&1.17	&11.39	&3.96	&9.46\\
15&2.41	&1.39	&8.23	&0.32	&1.39	&9.36	&2.46	&7.33\\
20&2.73	&1.93	&4.97	&-	&-	&8.66	&1.93	&7.65\\
        \bottomrule
    \end{tabular}
    \caption{Percentages of answering ``I don't know'' across models and settings.}
    \label{tab:idk_rate}
\end{table*}
\section{Annotation details}
\label{sec:app-anno}

For organism species selection in query construction, we aim to ensure that the chosen species are relatively under-represented in textual resources.To achieve that, we filter them based on the length of their Wikipedia summary and description sections. Well-known species typically have multiple sections (e.g., habitats, behaviour, etc.), indicating extensive documentation. We therefore include only those species whose summary and description account for no more than 50\% length of the entire Wikipedia article, as an attempt on minimizing reliance on prior knowledge: widely recognized species may have detailed descriptions on visual features across LLM pre-training corpora, even if these details are not explicitly mentioned in Wikipedia.

For the query generation, we mentioned that we have gone through a human annotator originated refined filtering and rewriting process. The prompt template is shown in Table \ref{tab:prompt_qgen}, and following shows the example prototype queries generated with the sample prompt:

\begin{figure}[t]
    \centering
    \includegraphics[width=0.9\linewidth]{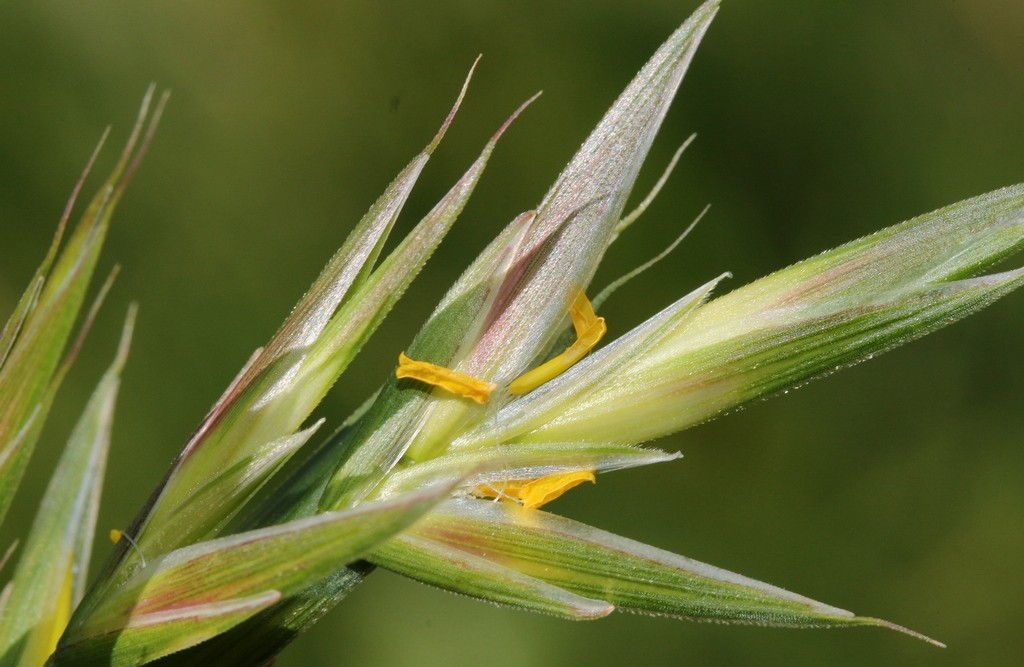}
    \caption{Example image of Bromus catharticus flower anther. Owned by original uploader at www.inaturalist.org, shared under CC-BY-NC license.}
    \label{fig:sample_qgen}
\end{figure}

\begin{enumerate}
    \item Does rescuegrass (scientific name: Bromus catharticus) have a visible ligule when the leaf sheath is gently parted, and if so, what is its shape? (Explanation: The ligule is a thin membrane at the junction of leaf blade and sheath, typically hidden in standard photos) (Feature: shape of ligule) \textit{Dropped. Too detailed and hard to distinguish}

    \item During the flowering stage of rescuegrass (scientific name: Bromus catharticus), what color do the exposed anthers display? (Explanation: The anthers are only briefly visible and require a closer or angled view) (Feature: color of anthers) \textit{Passed. Annotator found a image of flower anther (Figure \ref{fig:sample_qgen})  and confirm it is rare image.}

    \item Is there any distinct pattern or texture on the backside of the spikelets of rescuegrass (scientific name: Bromus catharticus), observable only when the spikelets are lifted or spread apart? (Feature: pattern on backside of spikelets) \textit{Dropped. Viewing the spikelets image, there is no ``backside'' on spikelets as they are symmetric.}

    \item When the plant is fully mature, how does the collar region (the area where the leaf blade meets the sheath) of rescuegrass (scientific name: Bromus catharticus) appear from the underside view? (Explanation: This area may be covered by overlapping leaves in typical photos) (Feature: shape/appearance of collar region) \textit{Dropped. Query too obscure and likely no clue image can be found.}
\end{enumerate}

Human annotator carefully check the candidate prototype queries by browsing all available material online with the scientific name as search keyword, as well as briefly scan through the image corpus by viewing the image thumbnails.

\begin{figure*}[b]
    \centering
    \includegraphics[width=0.9\textwidth]{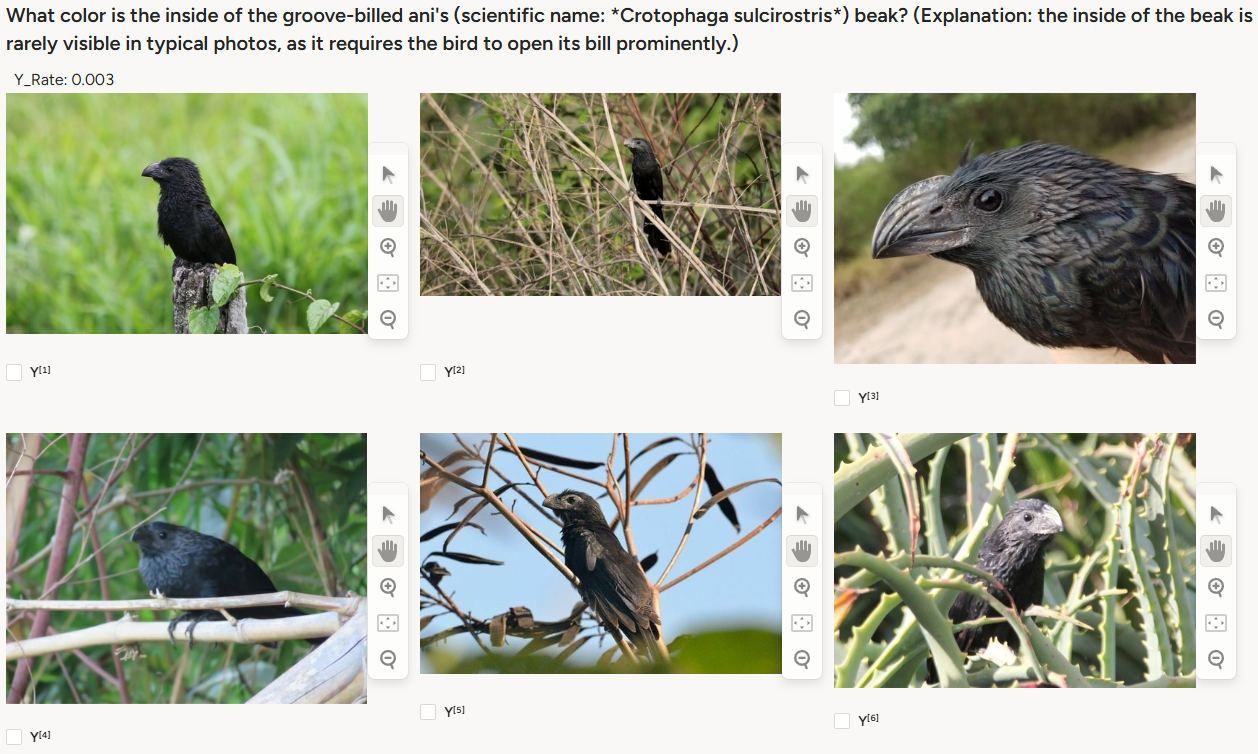}
    \caption{Annotation interface for image labelling.}
    \label{fig:anno_interface}
\end{figure*}

\begin{figure*}[b]
    \centering
    \includegraphics[width=0.9\textwidth]{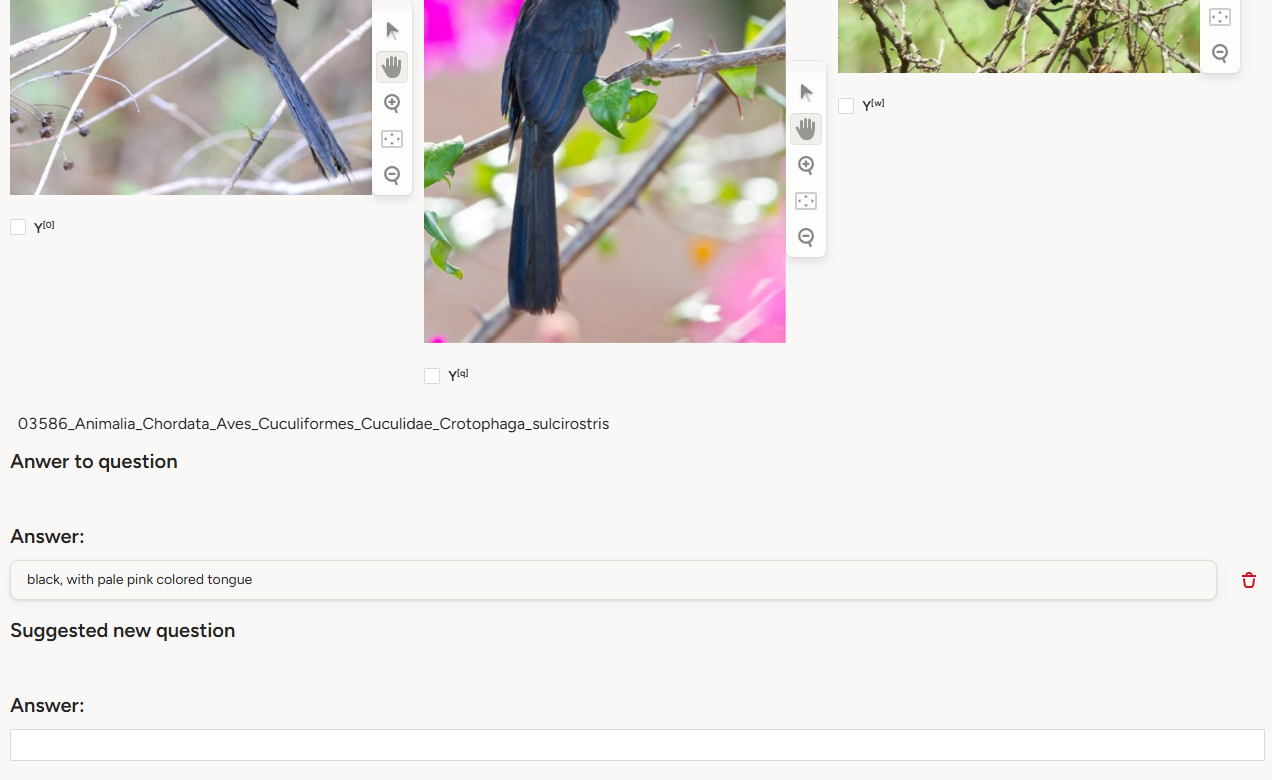}
    \caption{Annotation interface for image label and query answer.}
    \label{fig:anno_interface_ans}
\end{figure*}

After query generation and rewriting, and the coarse-level filtering described in Section 4.2, we dispatch the validated queries to the next round of human annotation of image labels.
Figure \ref{fig:anno_interface} shows the annotation interface for image labelling.\footnote{Similar interface is used for query filtering and rewriting annotator to browse, except the image matrix is more dense and displaying only thumbnail sized images.} Annotators are required to only label ``Y'', to indicate that this image contain the queried feature and is a clue to the query. Leaving the checkbox blank will be considered as negative label. This design aims to save annotators' effort, as clue images are assumed to be the minority. The rough clue image rate predicted by MLLM in the previous step is also provided (``Y\_Rate: 0.003'' after the question text) as reference for annotators to estimate the query difficulty. In the meanwhile, annotators provide query answers upon finding a clue image, as shown in Figure \ref{fig:anno_interface_ans}. Annotators are allowed to skip for a query when finding more than 9 clues after the first 27 images, indicating this query likely fail the clue image threshold. The order of images are randomly shuffled.

\begin{table}[t]
    \centering
    \begin{tabular}{c|lll}
    \toprule
    Human\textbackslash LLM   & True &Partial &False\\
    \midrule
         True   &86 &1  &3  \\
         Partial    &6  &14 &2\\
         False  &6  &17 &65\\
     \bottomrule
    \end{tabular}
    \caption{Confusion matrix of human evaluation v.s. LLM-as-Judge. We consider ``partially correct'' as wildcard match, and the macro-$F_1$ score is 0.92.}
    \label{tab:confusion}
\end{table}

\section{Prompt Templates}
\label{sec:app_prompt}

Table \ref{tab:prompt_qgen}, \ref{tab:prompt_qa}, \ref{tab:prompt_eval} list the prompt templates used through out our experiments.

\begin{table*}[t]
\centering
\small
\begin{tabular}{p{0.97\textwidth}}
\toprule[2pt]
\textbf{Prompt Template for Query Generation}\\ \midrule
Given an organism's scientific taxonomy name and common name (if available), please generate some questions regarding its visual feature, like color, shape, pattern, texture, etc. Please limit the question asked to be regarding features that may not be usually found in photos. This can be regarding the following but not limited to: uncommon-to-see parts of animals or plants / parts that can only be viewed at specific posture or require photo taken at specific angle / less commonly seen stage of life cycle of incect, etc. (e.g. ``What is the color of paw pad of giant panda'', photos showing the paw pad of giant panda would be only a small portion; ``What pattern is on the back of brown peacock's larva?'', asking for the larva form of this butterfly.) Questions regarding non-visual feature should not be asked. (e.g., ``how long can giant panda live'', ``what does puma eat'' are not suitable questions, as such information cannot be extracted from a photo of that animal). Please avoid questions regarding exact length/area (of tails, ears, leaf, etc.), as such attribute is hard to be estimated from photo. Please avoid questions regarding eggs, nests, nets, etc., and focus on the organism's body. Please avoid generating multiple paraphrased questions regarding a same feature or body part. Please avoid questions comparing another species / form, but you can compare within the same individual (e.g. Does the upper beak or lower beak of XXX bird have darker color?).\\Except these constraints, try to generate ``interesting'' questions: for example, photos showing underside of leaf may be rare, but asking what color is underside of leaf is not interesting -- it's green anyway. Hence after generating each question, try to check if the questions is ``interesting'' or not. Please avoid asking for features that ``technically'' appear on a photo but hard to see (e.g., ``What color is the tiny hair on the surface of XYZ plant's stem?'' Any photo displays the stem will show the tiny hair, but it needs to be zoomed in / photo taken at very close, to see. The desired questions is regarding features that is hidden in some photos, not features that appear in every photo but simply too small to see).\\A likely related paragraph from Wikipedia is provided for you to better understand the organism. Please avoid question regarding visual features explicitly mentioned in the Wikipedia paragraphs. (e.g., paragraph: ``The seeds are colored in red'' -$>$ Don't ask ``What color is the seed''; paragraph: ``The plant have symbolic seed pods'' (but doesn't mention its color) -$>$ You may ask ``What shape/color is the seed pod'')\\Following are format instructions. You don't need to give answer to the questions.Please generate questions by asking the common name and add scientific name in brackets, and indicate the questioned visual feature after the question sentence, e.g.: ``What color is *** of \{common\_name\} (scientific name: \{scientific\_name\})? (Feature: color of ***)''; if there is no common name, simply ask with the scientific name. If explanation to the question is needed (such as explain a technical term), do not start a new line, append the explanation after the question enclosed by small brackets ().\\\\
Following are a few examples:
\\Scientific name: Diastictis fracturalis\\Common name: fractured western snout moth\\Question:\\1. What pattern, if any, can be observed on fractured western snout moth's (scientific name: Diastictis fracturalis) underside of wing during flying? (Explanation: the moth's wing usually cover its whole abdomen when resting, and underside will be less commonly appear in photos) (Feature: pattern on underside of wing during flying)\\2. What is the common color of larva fractured western snout moth (scientific name: Diastictis fracturalis)? (Feature: color of larva)?
\\\\Scientific name: Tortula muralis\\Common name: wall-screw moss\\Question:\\1. What is the common color of tip of leaf of wall-screw moss (scientific name: Tortula muralis)? (Explanation: The moss can be distinguished from similar mosses by its ``air-pointed'' leaves.) (Feature: color of leaf tip)
\\\\Scientific name: Ptiliogonys cinereus\\Common name: Gray Silky-flycatcher\\Question:\\1. What color are the undertail covert (short plumages that cover the long tail feather) of Gray Silky-flycatcher (scientific name: Ptiliogonys cinereus) when viewing from front? (Feature: color of undertail covert viewing from front)\\2. What pattern can be observed on the tail feather of Gray Silky-flycatcher (scientific name: Ptiliogonys cinereus) from dorsal (back) view? (Feature: pattern on tail feather from dorsal view)
... \textit{More ICL Examples omitted} ...
\\

End of examples.
\\\\Now, please generate questions for the given scientific name.\\
\textit{Example new query}\\
Scientific name: Bromus catharticus\\Common Name: Rescuegrass\\Wiki Paragraph: Bromus catharticus is a species of brome grass known by the common names rescuegrass, grazing brome, prairie grass, and Schrader's bromegrass. The specific epithet catharticus is Latin, meaning cathartic. The common name rescuegrass refers to the ability of the grass to provide forage after harsh droughts or severe winters. The grass has a diploid number of 42 ...... \textit{Omitted due to space. Actual prompt includes whole description passage.}\\Question:\\
\bottomrule
\end{tabular}
\caption{Prompt Template for Query Generation. Texts in \textit{italic} are not part of prompt.}
\label{tab:prompt_qgen}
\end{table*}


\begin{table*}
\centering
\begin{tabular}{p{0.97\linewidth}}

\toprule[2pt]
\textbf{Prompt Template for Main Question Answering Experiments of Different Settings}\\ \midrule[1.2pt]

\textbf{Zero-shot}\\
\midrule
Please answer the question regarding a visual feature of an organism (animal, plant, etc.). Please follow the answer format: ``Answer: \{answer\_text\}''\\\\Question: \\
\midrule[1.2pt]

\textbf{GT clue}\\
Please answer the question regarding a visual feature of an organism (animal, plant, etc.). You will be provided with a image regarding that organism, it is likely to contain the key information for answering the question. Please follow the answer format: ``Answer: \{answer\_text\}''\\\\Question:\\$<$image$>$\\
\midrule[1.2pt]

\textbf{Non-Clue}\\
Please answer the question regarding a visual feature of an organism (animal, plant, etc.). You will be provided with a image regarding that organism. Please follow the answer format: ``Answer: \{answer\_text\}''\\\\Question:\\$<$image$>$\\
\midrule[1.2pt]

\textbf{Multiple Images}\\
Please answer the question regarding a visual feature of an organism (animal, plant, etc.). You will be provided with several images, all of them are regarding that organism, but not all images contain the key information for answering the question. Only one to few images, or even none of them can be used to answer. Please follow the answer format: ``Answer: \{answer\_text\}''\\\\Question:\\ $<$image$>$ $<$image$>$ $<$image$>$...\\
\bottomrule
\end{tabular}
\caption{Prompt Template for Main Question Answering Experiments of Different Settings.}
\label{tab:prompt_qa}
\end{table*}


\begin{table*}
\centering
\small
\begin{tabular}{p{0.97\textwidth}}

\toprule[2pt]
\textbf{Prompt Template For Automatic Evaluation}\\ \midrule

Please evaluate the answer to a question, score from 0 to 1. The reference answer is provided, and the reference is usually short phrases or a single keyword. If the student answer is containing the keywords or similar expressions (including similar or close color/pattern), without any additional guessed information, it is full correct. Similar or close color/pattern includes but not limited to the following cases: pale or light color can appear to be yellowish/greyish under different light condition; dark colors like dark brown, dark grey, dark purple can appear close to each other, and may appear as black as well; stripe pattern can appear as band or ring, dotted pattern can  etc. If the student answer have missed some important part in the reference answer, please assign partial score. The reference answer can be in the form of a Python list, in this case, any one of the list item is correct. \\If student answer contain irrelevant information not related to question, mark it with ``Redundant'', but it does not affect score if related parts are correct. (e.g. Question: what shape are leaves of XYZ plant, Student Answer: shape xxx, color yyy, color is Redundant answer)\\If student answer contain features not listed in reference answer, deduct 0.5 score and mark it with ``Likely Hallucination''. (e.g., Reference Answer: black and white. Student Answer: black white, with yellow dots, ``Yellow dots'' is not mentioned in reference). The reference answer sometimes contains additional information not asked in question, usually enclosed by brackets (), to help verifying hallucinations (e.g.: ``Shape is xxx (color is yyy)''). Not mentioning additional information in answer is not considered wrong. For yes/no question, reference may contain explanations on why giving yes/no, but it is not necessary for student to answer the explanation; however, if student explains and differs with reference, it is considered as hallucination. Answering ``I don't know'', ``Not enough information'' or similar is considered wrong, and please mark it with ``No Answer''.\\\\Format Instructions: Separate the remarks with score using ``$\vert$'', that is, use the syntax of: ``Score: {score} $\vert$ Likely Hallucination'', ``Score: {score}'', ``Score: {score} $\vert$ Likely Hallucination $\vert$ Redundant'', ``Score: 0 $\vert$ No Answer ''. If any explanation on why giving the score is needed, do not start a new line and append after remark with brackets, e.g. ``Score: {score} $\vert$ Redundant $\vert$ (Explanation: abc)''.\\\\Following are few examples:\\\\
Question: Is there any specific color marking around the eyes of a semipalmated plover (scientific name: Charadrius semipalmatus)?\\Reference Answer: black eye-round feather, white stripe above eyes. (sometimes connected to the white forehead)\\\\Student Answer: Yes, the bird has a distinctive black line that runs through the eye, which is a key identifying feature.\\Score: 0 $\vert$ Likely Hallucination\\\\Student Answer: They have a black vertical band in front of the eye, a white band above the eye, and a single black band that wraps partially around the eye, creating a partial ``mask'' appearance.\\Score: 1\\Student Answer: Yes, the semipalmated plover has a distinctive black/dark ring around its eye, surrounded by a bright white ring or patch\\Score: 0.5 $\vert$ Likely Hallucination (Explanation: not white ring, but only a line above the eye)\\\\\\Question: What is the typical color of the antennae of Harris's checkerspot butterfly (scientific name: Chlosyne harrisii)?\\Reference Answer: alternating black and white band, with yellow on the tip\\\\Student Answer: The antennae of Harris's checkerspot butterfly are black with orange-tipped clubs.\\Score: 0.5 (Explanation: not mentioning black and white)\\\\Student Answer: The typical color of the antennae of Harris's checkerspot butterfly is black with white spots.\\Score: 0.5 $\vert$ Likely Hallucination (Explanation: not white spot but band. Not mentioning the tip)\\\\\\Question: Are the leaves of burro-weed (scientific name: Ambrosia dumosa) usually covered in small hairs?\\Reference Answer: yes\\\\Student Answer: Yes, the leaves of burro-weed (Ambrosia dumosa) are typically covered in small hairs, giving them a grayish or whitish-green appearance.\\Score: 1 $\vert$ Redundant\\\textit{... More ICL examples omitted...}\\
\\Now, Score the following question:\\
\bottomrule
\end{tabular}
\caption{Prompt Template For Automatic Evaluation. Note that for evaluation, no images is provided, and it becomes pure-text open-ended QA evaluation. Texts in \textit{italic} are not part of prompt.}
\label{tab:prompt_eval}
\end{table*}

\section{Detailed Results}
\label{sec:app_results}

This section contains detailed experimental results. 
The breakdown results for main experiments can be found in Table~\ref{tab:baseline_detail}, \ref{tab:1ink_detail}.
We also include ROUGE-1 Recall scores\footnote{Implemented with https://pypi.org/project/rouge/, version 1.0.1} for readers who are interested in Table~\ref{tab:baseline_rouge} and \ref{tab:1ink_rouge}. However, ROUGE relies on strict word matching, the results for open-ended QA can vary significantly due to the different generation styles across models.


\begin{table*}[t]
    \centering
    \begin{tabular}{l|ccccc|ccc}
    \toprule
         Setting&Phi4-MM	&Qwen2.5VL	&InternVL3	&Pixtral	&Llama3.2-V	&GPT-4o	&Gemini	&Claude\\
         \midrule
        Zeroshot   &35.16	&38.90	&39.17	&41.71	&32.35	&53.74	&60.43	&54.28\\
        Non-clue&
34.76	&30.08	&29.28	&42.74	&40.37	&14.97	&17.11	&21.39\\
         \midrule
\multirow{5}{*}{GT clue}&
45.05	&42.51	&42.11	&46.66	&46.26	&62.17	&62.17	&55.35\\
&44.92	&38.50	&41.44	&47.86	&45.99	&56.95	&61.76	&60.56\\
&45.94	&43.58	&43.32	&47.73	&47.46	&60.16	&63.10	&55.88\\
&46.79	&43.72	&44.92	&45.19	&50.40	&60.29	&64.57	&55.88\\
&42.51	&40.64	&46.66	&48.13	&48.93	&59.49	&62.79	&56.28\vspace{0.3em}\\
\midrule
AVG&45.04	&41.79	&43.69	&47.11	&47.81	&59.81	&62.88	&56.79\\
STD&1.604	&2.213	&2.123	&1.212	&1.857	&1.884	&1.081	&2.133\\

\bottomrule
    \end{tabular}
    \caption{Baselines detailed results}
    \label{tab:baseline_detail}
\end{table*}

\begin{table*}[t]
    \centering
    \begin{tabular}{l|ccccc|ccc}
    \toprule
         Setting&Phi4-MM	&Qwen2.5VL	&InternVL3	&Pixtral	&Llama3.2-V	&GPT-4o	&Gemini	&Claude\\
         \midrule
        Zeroshot&21.46	&27.06	&25.66	&25.88	&16.63	&24.00	&26.47	&35.38\\
        Non-clue    &13.96	&12.33	&19.81	&23.52	&3.96	&13.76	&16.01	&24.14\\

         \midrule
\multirow{5}{*}{GT clue}
&18.79	&27.78	&20.64	&28.11	&15.55	&25.25	&24.34	&32.36\\
&19.29	&25.89	&21.16	&26.74	&15.73	&26.51	&22.44	&33.88\\
&18.74	&27.43	&21.36	&27.68	&16.18	&25.64	&22.94	&33.45\\
&19.70	&27.49	&22.71	&26.64	&16.27	&24.26	&24.15	&32.37\\
&18.91	&26.89	&24.19	&26.48	&14.53	&26.67	&24.38	&33.46\vspace{0.3em}\\
\midrule
AVG&19.09	&27.10	&22.01	&27.13	&15.65	&25.67	&23.65	&33.10\\
STD&0.405	&0.747	&1.437	&0.721	&0.696	&0.984	&0.898	&0.697\\
\midrule
\multicolumn{9}{c}{\textit{Top-K Retrieval-Augmented Generation}}\\
\midrule
k=1&16.03	&13.58	&21.71	&23.58	&4.54	&14.48	&20.57	&28.22\\
k=3&16.92	&7.95	&20.66	&22.32	&7.37	&19.60	&24.50	&32.78\\
k=5&14.99	&11.45	&18.98	&21.92	&5.46	&22.59	&25.67	&34.04\\
k=7&16.15	&12.81	&17.62	&21.51	&6.11	&24.19	&27.92	&34.73\\
k=10&16.61	&13.88	&18.40	&22.49	&6.73	&23.49	&28.17	&35.18\\
k=15&16.55	&13.55	&16.94	&24.60	&5.72	&23.74	&31.12	&35.43\\
k=20&16.75	&13.51	&14.93	&-	&-	&25.50	&30.30	&37.95\\
\bottomrule
    \end{tabular}
    \caption{Baselines and Top-K RAG ROUGE scores for reference. Please note that ROUGE score fluctuates largely across models, as the generated text style largely affect the matching with ground truth.}
    \label{tab:baseline_rouge}
\end{table*}

\begin{table*}[t]
    \centering
    \begin{tabular}{l|ccccc|ccc}
\toprule
1 in K&	Phi4-MM	&Qwen2.5VL	&InternVL3	&Pixtral	&Llama3.2-V	&GPT-4o	&Gemini	&Claude\\
\midrule
\multirow{5}{*}{k=3}&	41.04	&46.93	&39.44	&48.26	&46.26	&47.99	&59.09	&48.26\\
&43.54	&47.46	&41.67	&46.12	&46.75	&50.22	&60.29	&48.80\\
&40.78	&45.05	&41.18	&48.80	&48.53	&47.33	&58.82	&47.46\\
&41.98	&46.26	&43.85	&45.86	&47.73	&50.27	&60.29	&47.33\\
&42.25	&48.53	&39.17	&46.52	&44.39	&48.93	&59.36	&48.40\\
\midrule
AVG&41.92	&46.85	&41.06	&47.11	&46.73	&48.95	&59.57	&48.05\\
STD&1.097	&1.303	&1.895	&1.329	&1.576	&1.314	&0.684	&0.632\\
\midrule
\multirow{5}{*}{k=5}&43.45	&48.13	&37.97	&46.39	&47.59	&54.41	&61.50	&50.00\\
&39.30	&48.53	&40.37	&47.59	&45.05	&51.28	&59.49	&49.06\\
&38.50	&47.55	&36.32	&46.52	&43.72	&54.28	&62.57	&49.06\\
&39.44	&49.06	&38.24	&46.66	&45.86	&52.67	&59.36	&50.40\\
&39.97	&46.93	&39.30	&46.26	&43.72	&53.48	&60.56	&50.80\\
\midrule
AVG&40.13	&48.04	&38.44	&46.68	&45.19	&53.22	&60.70	&49.86\\
STD&1.928	&0.831	&1.518	&0.528	&1.624	&1.292	&1.362	&0.787\\
\midrule
\multirow{5}{*}{k=7}&40.37	&48.80	&40.64	&46.39	&43.32	&56.68	&61.93	&51.20\\
&41.04	&47.06	&36.36	&47.59	&42.51	&56.37	&61.90	&52.81\\
&39.57	&47.46	&40.64	&46.52	&46.79	&54.41	&61.36	&50.48\\
&40.64	&48.26	&41.44	&46.66	&42.38	&51.34	&61.76	&50.94\\
&42.38	&47.33	&39.71	&46.26	&45.99	&58.02	&60.70	&49.60\\
\midrule
AVG&40.80	&47.78	&39.76	&46.68	&44.20	&55.36	&61.53	&51.01\\
STD&1.034	&0.724	&1.996	&0.528	&2.053	&2.593	&0.517	&1.178\\
\midrule
\multirow{5}{*}{k=10}&40.24	&45.19	&37.43	&44.65	&41.44	&53.61	&62.70	&50.27\\
&41.00	&47.06	&39.44	&46.12	&37.17	&57.49	&60.56	&51.07\\
&37.97	&48.66	&38.64	&44.79	&42.38	&56.23	&63.90	&53.88\\
&39.57	&49.06	&38.50	&44.61	&42.91	&55.75	&63.64	&53.21\\
&36.63	&47.19	&38.50	&43.98	&45.59	&58.16	&62.70	&50.00\\
\midrule
AVG&40.80	&47.43	&38.50	&44.83	&41.90	&56.25	&62.70	&51.69\\
STD&1.034	&1.531	&0.716	&0.786	&3.060	&1.761	&1.314	&1.758\\
\midrule
\multirow{5}{*}{k=15}&41.31	&51.34	&39.30	&44.92	&43.32	&57.62	&64.17	&52.01\\
&39.57	&49.73	&37.83	&43.72	&39.17	&55.88	&64.71	&51.20\\
&39.71	&51.07	&38.90	&43.58	&38.50	&55.08	&63.90	&54.68\\
&42.25	&47.59	&39.17	&43.54	&42.11	&57.49	&62.37	&54.81\\
&39.17	&48.13	&40.51	&43.98	&42.51	&55.75	&64.88	&50.67\\
\midrule
AVG&39.08	&49.57	&39.14	&43.95	&41.12	&56.36	&64.01	&52.67\\
STD&1.768	&1.688	&0.959	&0.570	&2.146	&1.130	&0.997	&1.950\\
\midrule
\multirow{5}{*}{k=20}&41.58	&47.33	&39.30	&-	&-	&56.95	&65.37	&52.54\\
&39.84	&49.33	&37.03	&-	&-	&57.22	&64.57	&50.13\\
&41.18	&50.94	&39.44	&-	&-	&54.81	&62.03	&53.21\\
&39.71	&47.15	&42.78	&-	&-	&55.48	&62.43	&51.60\\
&41.84	&47.99	&36.90	&-	&-	&58.02	&62.97	&53.21\\
\midrule
AVG&37.25	&48.55	&39.09	&-	&-	&56.50	&63.47	&52.14\\
STD&0.992	&1.588	&2.389	&-	&-	&1.316	&1.434	&1.302\\
\bottomrule
    \end{tabular}
    \caption{1-in-K detailed results}
    \label{tab:1ink_detail}
\end{table*}

\begin{table*}[t]
    \centering
    \begin{tabular}{l|ccccc|ccc}
\toprule
1 in K&	Phi4-MM	&Qwen2.5VL	&InternVL3	&Pixtral	&Llama3.2-V	&GPT-4o	&Gemini	&Claude\\
\midrule
\multirow{5}{*}{k=3}&16.93	&7.05	&19.39	&24.89	&6.12	&20.61	&28.82	&35.80\\
&16.88	&7.43	&21.30	&25.07	&6.23	&22.40	&29.24	&35.15\\
&16.55	&6.37	&20.01	&25.09	&6.70	&20.58	&27.82	&32.25\\
&17.34	&7.24	&22.95	&23.82	&5.70	&20.96	&28.97	&33.58\\
&15.25	&7.54	&19.71	&24.59	&5.66	&21.69	&29.18	&35.05\\
\midrule
AVG&16.59	&7.13	&20.67	&24.69	&6.08	&21.25	&28.81	&34.37\\
STD&0.800	&0.462	&1.466	&0.527	&0.427	&0.784	&0.576	&1.435\\
\midrule
\multirow{5}{*}{k=5}&17.55	&9.72	&19.33	&24.35	&5.63	&24.47	&27.78	&35.70\\
&15.90	&10.50	&18.26	&24.85	&7.20	&22.47	&29.19	&36.12\\
&15.54	&11.02	&15.34	&24.03	&6.12	&21.93	&30.06	&33.99\\
&16.46	&11.55	&20.07	&23.60	&5.90	&22.01	&28.95	&34.39\\
&16.22	&10.36	&19.00	&23.37	&5.86	&23.01	&28.13	&34.48\\
\midrule
AVG&16.33	&10.63	&18.40	&24.04	&6.14	&22.78	&28.82	&34.94\\
STD&0.763	&0.692	&1.830	&0.591	&0.616	&1.039	&0.901	&0.920\\
\midrule
\multirow{5}{*}{k=7}&14.37	&12.70	&17.83	&23.93	&7.25	&24.52	&29.94	&34.47\\
&14.28	&12.09	&16.69	&24.10	&6.11	&23.09	&28.13	&34.93\\
&15.18	&10.95	&18.25	&23.60	&6.09	&24.05	&29.74	&32.26\\
&15.65	&11.51	&17.68	&23.81	&6.04	&23.47	&30.29	&35.47\\
&16.17	&11.99	&18.22	&22.65	&5.78	&23.77	&31.00	&35.81\\
\midrule
AVG&15.13	&11.85	&17.73	&23.62	&6.25	&23.78	&29.82	&34.59\\
STD&0.815	&0.657	&0.633	&0.571	&0.572	&0.546	&1.060	&1.398\\
\midrule
\multirow{5}{*}{k=10}&16.04	&12.20	&15.97	&22.10	&6.06	&21.83	&31.15	&34.62\\
&14.90	&12.62	&15.92	&22.78	&6.11	&22.11	&30.38	&35.07\\
&15.24	&13.65	&16.39	&23.42	&5.64	&24.25	&30.47	&35.48\\
&15.55	&12.46	&16.59	&23.37	&5.39	&22.95	&32.16	&34.23\\
&15.56	&12.26	&17.49	&23.85	&7.08	&25.03	&29.40	&33.65\\
\midrule
AVG&15.46	&12.64	&16.47	&23.10	&6.06	&23.23	&30.71	&34.61\\
STD&0.423	&0.590	&0.635	&0.678	&0.646	&1.376	&1.022	&0.713\\
\midrule
\multirow{5}{*}{k=15}&15.43	&12.39	&17.70	&26.02	&5.51	&25.42	&31.44	&34.84\\
&15.26	&13.18	&16.60	&25.30	&7.23	&24.87	&31.85	&35.15\\
&15.22	&12.76	&14.40	&25.35	&5.49	&24.45	&30.32	&34.60\\
&15.32	&12.57	&15.20	&25.10	&5.72	&24.37	&31.54	&35.83\\
&14.91	&12.79	&16.12	&26.04	&5.76	&24.90	&32.22	&35.69\\
\midrule
AVG&15.23	&12.74	&16.00	&25.56	&5.94	&24.80	&31.47	&35.22\\
STD&0.195	&0.295	&1.271	&0.437	&0.730	&0.420	&0.713	&0.531\\
\midrule
\multirow{5}{*}{k=20}&16.24	&12.60	&15.07	&-	&-	&24.29	&31.76	&36.16\\
&15.03	&13.46	&14.38	&-	&-	&23.72	&31.80	&34.58\\
&16.00	&13.64	&15.57	&-	&-	&23.43	&30.94	&36.53\\
&15.85	&13.42	&15.13	&-	&-	&24.37	&31.47	&34.87\\
&15.40	&13.03	&14.35	&-	&-	&24.09	&31.84	&35.40\\
\midrule
AVG&15.70	&13.23	&14.90	&-	&-	&23.98	&31.56	&35.51\\
STD&0.486	&0.417	&0.525	&-	&-	&0.397	&0.377	&0.829\\

\bottomrule
    \end{tabular}
    \caption{1-in-K ROUGE scores for reference. Please note that ROUGE score fluctuates largely across models, as the generated text style largely affect the matching with ground truth.}
    \label{tab:1ink_rouge}
\end{table*}
\section{Qualitative Error Case Analysis}

Providing the ground-truth clue images effectively makes the problem a standard VQA task, which is considered to be ``solved'' for the state-of-the-art proprietary models. However, our experiment results shown that their improvements over zero-shot baseline is marginal. We present a few typical error cases from proprietary models under GT clue setting in Table~\ref{tab:error_case}.

\begin{table*}[t]
    \centering
    \begin{tabular}{p{0.95\textwidth}}
    \toprule
    \includegraphics[width=0.5\textwidth]{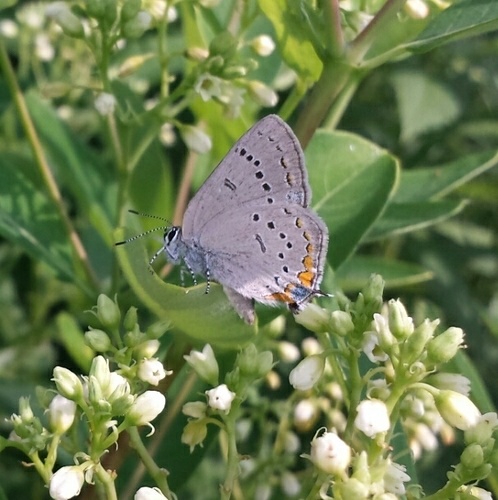}\\
    Question: When resting, Acadian Hairstreak (scientific name: Satyrium acadica) usually fold its wings and the abodmen is hidden; what color is its abdomen?\\
    Ground Truth Answer: [white, light grey]\\
    Gemini's Answer: In the provided image, the abdomen of the Acadian Hairstreak is obscured by its folded wings, making its color not visible.\\\\
    \midrule
    \\\\
    \includegraphics[width=0.5\textwidth]{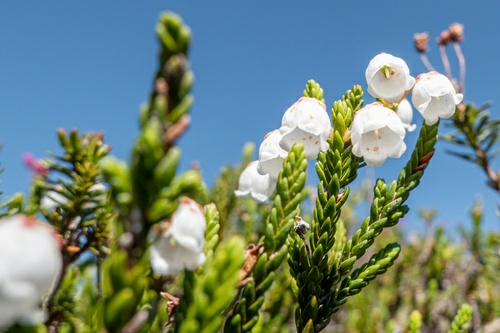}\\
    Question: What color is the stamen tip inside the bell-shaped flower of western moss-heather (scientific name: Cassiope mertensiana)? (Explanation: opening of the flower usually points downward)\\
    Ground Truth Answer: light yellow\\
    Gemini's Answer: Red\\
    \bottomrule

    \end{tabular}
    \caption{Sample cases that Gemini failed to answer even with ground-truth clue image. Fine-grained details in the image of area of few pixels remain challenging even for the state-of-the art proprietary models.}
    \label{tab:error_case}
\end{table*}

\end{document}